%% file: neurips_2026.tex
\documentclass{article}

\PassOptionsToPackage{numbers,compress}{natbib}

\usepackage[preprint]{neurips_2026}

\usepackage[utf8]{inputenc}
\usepackage[T1]{fontenc}
\usepackage{hyperref}
\usepackage{url}
\usepackage{booktabs}
\usepackage{amsfonts}
\usepackage{amsmath}
\usepackage{amssymb}
\usepackage{nicefrac}
\usepackage{microtype}
\usepackage{xcolor}
\usepackage{graphicx}
\usepackage{tikz}
\usepackage{threeparttable}
\usepackage{colortbl}
\usepackage{multirow}
\usepackage{wrapfig}
\usepackage{xspace}
\usepackage{subcaption}
\usepackage{pifont}
\usepackage[most]{tcolorbox}
\usepackage{fontawesome5}
\usepackage{placeins}

\definecolor{deepblue}{RGB}{0,51,153}
\definecolor{darkgreen}{HTML}{008000}
\definecolor{lightblue}{HTML}{bde0fe}
\graphicspath{{./}{figs/}}

\newcommand{\cmark}{\ding{51}}
\newcommand{\xmark}{\ding{55}}

\newcommand{\ignore}[1]{}

\makeatletter
\DeclareRobustCommand\onedot{\futurelet\@let@token\@onedot}
\def\@onedot{\ifx\@let@token.\else.\null\fi\xspace}
\def\eg{\emph{e.g}\onedot}

\def\ie{\emph{i.e}\onedot}

\def\etc{\emph{etc}\onedot}

\makeatother

\newtcolorbox{fancybox}[2][blue]{ 
  enhanced,
  breakable,
  colframe=#1!70!black,
  colback=#1!5,
  coltitle=black,
  colbacktitle=#1!25,
  fonttitle=\bfseries,
  title=#2,
  boxed title style={
    colframe=#1!70!black,
    colback=#1!25,
    sharp corners,
    size=small,
    interior style={top color=#1!25, bottom color=#1!25}
  },
  top=2mm, bottom=2mm, left=4mm, right=4mm,
  boxrule=0.8pt,
  arc=4mm,
}

\title{Design Space for Scaling Continuous Memory in Multimodal Computer-Use Agents}

\author{%
  Wenyi Wu$^{1}$ \quad
  Kun Zhou$^{1}$\thanks{Corresponding author. Email: \texttt{franciskunzhou@gmail.com}.} \quad
  Sibo Zhu$^{1}$ \quad
  Ruoxin Yuan$^{2}$\thanks{This work was conducted during the author's internship at UCSD.}\\
  \textbf{Vivian Yu$^{1}$ \quad Stephen Wang$^{3}$ \quad Zhiting Hu$^{1}$ \quad Biwei Huang$^{1}$}\\[0.4em]
  $^{1}$University of California, San Diego\\
  $^{2}$Fudan University \qquad $^{3}$Abel AI\\
  \texttt{\{wew058,siz030\}@ucsd.edu}
}

\begin{document}

\maketitle

\begin{abstract}
To better generalize across unfamiliar scenarios and tasks, equipping Computer-Use Agents with external memory has emerged as an important topic. Existing methods convert past trajectories into lengthy text sequences to compose the memory bank, which increases the context length required for memory augmentation and can omit key visual cues (\eg widget size and position). To reduce the cost while preserving visual information, we leverage the continuous memory that encodes each trajectory into fixed-length continuous embeddings using the VLM itself as an encoder. We systematically study the design space of continuous memory along four dimensions: the number of in-context samples, memory bank size, backbone model size, and the number of continuous embeddings. Within the evaluated ranges, we observe positive empirical scaling trends as we increase the number of in-context samples, the memory bank size, and the backbone model size. To take advantage of these trends, we create a low-cost data synthesis pipeline to collect 188,451 trajectories. By fine-tuning only 1.2\% of parameters in the memory encoder with 1,500 samples, our memory-augmented agent consistently improves success rates under long horizons and distribution shifts. Notably, the Qwen-VL series of models, ranging from 2B to 32B, equipped with continuous memory achieve comparable performance to state-of-the-art closed-source models (\eg GPT-4o, Gemini-2.5-Pro).
\end{abstract}

\section{Introduction}

Recent advances in visual grounding and training techniques for vision-language models (VLMs)~\cite{mckinzie2024mm1methodsanalysis, wu2025} have catalyzed rapid progress in multimodal Computer-Use Agents (CUAs)~\cite{zhu2025oagentsempiricalstudybuilding, tongyidr},
which can operate across websites and desktop software to solve multi-step planning tasks such as information search and computer control. However, real-world tasks are often more unpredictable and long-horizon.
Existing CUAs generally do not perform well in unfamiliar visual layouts, iconography, and unseen functionalities~\cite{androidcontrol, guiodyssey}.
By contrast, humans exhibit strong robustness across diverse interfaces. Using accumulated episodic experiences and encountered interface states from human memory, we can easily learn to use new websites and apps, and accomplish novel tasks~\cite{Tulving2002}.

As a promising path toward better generalization, recent work equips CUAs with memories built from collected agent trajectories~\cite{fang2025memp, zhang2025universalretrievalmultimodaltrajectory}. Each trajectory is a sequence of screenshots and executed actions. Without compression, a single trajectory can span thousands, sometimes hundreds of thousands of tokens, inflating the inference cost and introducing irrelevant detail. To control cost, prior systems typically compress trajectories into text-based representations consisting of discrete language tokens. However, text cannot faithfully capture crucial visual cues in computer-use environments (\eg the precise size and position of GUI elements), which are often decisive for reliable execution.

To overcome these limitations, we aim to encode agent trajectories into continuous embeddings, instead of discrete text tokens. Concretely, we borrow the idea that uses a VLM as the encoder~\cite{wu2025}, and convert each multimodal CUA trajectory into a fixed-length sequence of 8 embeddings. In this way, we build a continuous memory bank consisting of trajectory representations with sharply reduced length while preserving fine-grained visual cues. At inference, relevant memory embeddings will be retrieved and then directly plugged into the VLM's embedding layer to inject new knowledge. This flexible integration alleviates the context-length bottleneck and supports scaling both the number of in-context items and the total memory size.

Based on the above design, we conduct a systematic investigation into the design space of scaling continuous memory across four dimensions: (i) in-context sample number, (ii) memory bank scale, (iii) backbone model scale, and (iv) continuous embedding number. Our findings reveal \textbf{positive empirical scaling trends within the evaluated ranges}: CUA performance improves as we scale the number of retrieved in-context samples, the memory bank size, and the backbone model size. The embedding count exhibits a separate, backbone-dependent optimum.

\begin{figure}
    \centering
    \includegraphics[width=\linewidth]{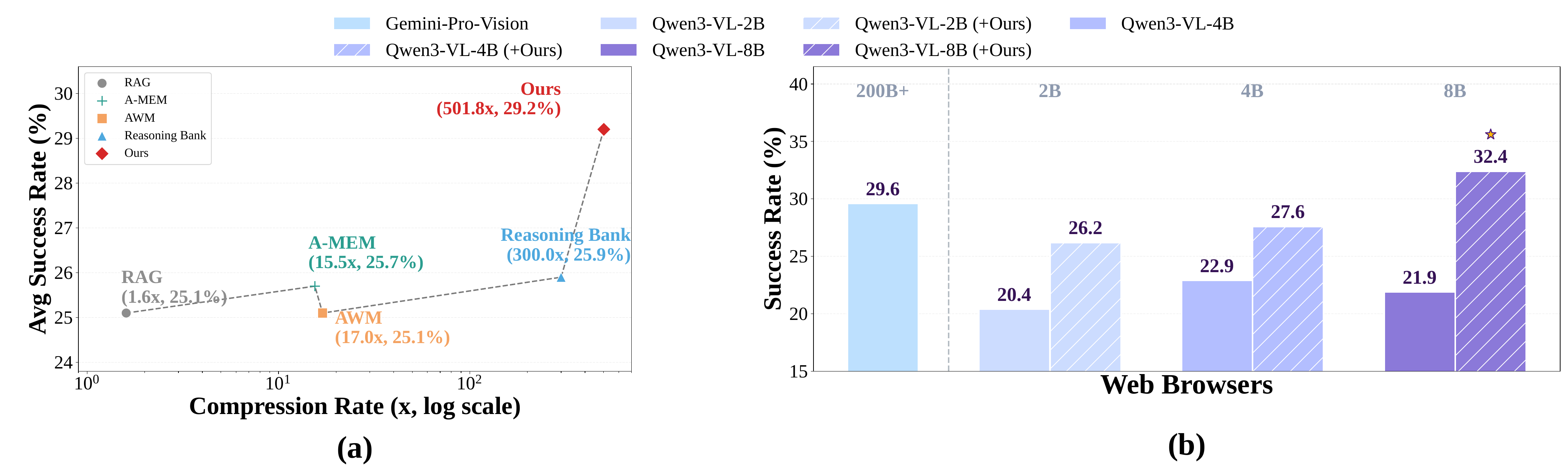}
    \caption{Comparison of different memory methods. (a) Using continuous memory (Ours) significantly outperforms other methods on Qwen2.5-VL-7B in both compression and success rates. (b) Applying continuous memory consistently boosts the backbone performance at different scales.}
    \label{fig:scaling_laws}
\end{figure}

To capitalize on these empirical trends, we seek to scale continuous memory by collecting far more agent trajectories. Yet high-quality traces are typically human-annotated, which are costly and hard to scale up. We therefore propose an \textbf{auto-scaling data flywheel} that autonomously discovers new environments, creates tasks, rolls out trajectories, and checks quality.
Using open-source VLMs in the pipeline, we collected a CUA dataset consisting of 188,451 trajectories using \$1,972 USD. 
To the best of our knowledge, it is the largest multimodal CUA trajectory dataset with fine-grained step-wise annotation, spanning over 35,000 environments.
Then, we leverage an existing CUA framework and augment different backbone models with the continuous memory built on the collected dataset.
Extensive experiments demonstrate that our memory leads to consistent improvements on real-world computer-use benchmarks and also yields lower inference cost, even under long-horizon and distribution shift scenarios.
As shown in Figure~\ref{fig:scaling_laws}, our method has a larger compression rate and better performance compared with other memory methods. The Qwen-VL series of models~\cite{yang2024qwen2.5,yang2025qwen3} ranging from 2B to 32B, once equipped with our memory method, can outperform state-of-the-art closed-source models (\eg GPT-4o~\cite{openai2024gpt4o}, Gemini-2.5-Pro~\cite{comanici2025gemini}), and show significant improvements in both web-browser and operating system (OS) scenarios.

\section{Continuous Memory for Computer-Use Agents}
\label{sec:comem}

\subsection{Computer-Use Agents}
\label{sec:prelim}

Following existing work~\cite{osworld,zheng2024seeact_multimodal_mind2web,bhathal2025websight,uitars}, 
we consider a multimodal Computer-Use Agent (CUA) that interacts with a digital environment 
(\eg web browser, desktop OS) through pixel-level observation 
and a set of platform-dependent input controls.
Given a natural language task instruction (\eg ``\textit{Find the address for the nearest Armageddon Shop}''), the agent perceives the world solely through screenshots and selects the next action from a unified action space $\mathcal{A}$ that includes a set of base actions common to all platforms such as $\textsc{CLICK}$, 
$\textsc{TYPE}$, $\textsc{SCROLL}$, \etc, and 
platform-specific extensions and tools. The full action space is detailed in Appendix~\ref{append:action_space}.
At time $t$, the agent receives an observation $o_t = \langle I_t \rangle$ where 
$I_t$ is the current screenshot, and outputs a structured action call 
$a_t \in \mathcal{A}$. The environment executes $a_t$ and returns the next 
screenshot $I_{t+1}$, yielding a trajectory $\tau=(o_t,a_t)_{t=1}^T$ up to a 
horizon $T_{\max}$.
A task is \emph{successful} if the final environment state satisfies the goal.

In this paper, we aim to equip CUAs with a continuous memory that consists of compressed embeddings of useful trajectories from different environments and tasks.
For each task, we first retrieve the top-$k$ relevant items from the memory, and then concatenate their continuous embeddings into the context.
At each step, the agent produces the next action conditioned on the current observation and memory items, and the memory-augmented policy is:
\begin{equation}
a_t \sim \pi_\theta\!\left(a \mid o_t,\, m_t\right),
\end{equation}
where $m_t$ is the concatenated continuous vectors of the retrieved memory items.

\subsection{Continuous Memory for CUA Trajectories}
\label{sec:memory design}

Leveraging the collected trajectories, we integrate a continuous memory into the CUA to enable effective knowledge reuse and transfer across different tasks and environments. 


\paragraph{Continuous Memory Encoder.}
We adapt CoMEM~\cite{wu2025} to the CUA setting, employing a memory encoder to compress each retrieved multimodal trajectory into a few continuous embeddings. The VLM encodes the entire trajectory, and learnable queries from the Q-Former~\cite{qformer} cross-attend to its last-layer representations, aggregating them into a fixed-length latent sequence regardless of trajectory length. These embeddings are injected into the agent in a plug-and-play manner by prepending them to the input embeddings, allowing the agent to attend to contextual memory without retraining. Thanks to the representational capacity of continuous embeddings, trajectories that often exceed 15,000 tokens can be compressed to as few as 8 vectors, fitting large external knowledge in limited context while preserving fine-grained visual cues critical for GUI navigation.

\paragraph{Memory Retrieval.}
At inference time, given the current observation $o_t$, we perform embedding-based retrieval to fetch the most relevant experiences. Concretely, a CLIP encoder~\cite{radford2021learning} maps screenshots and associated actions/queries from each trajectory to a set of embeddings, which are then pooled into a key vector. We index all keys with FAISS~\cite{douze2024faiss} and retrieve the top-$k$ nearest neighbors. The corresponding trajectories are converted into continuous embeddings by the memory encoder and then prepended to the agent's input, providing exemplar-driven guidance.

\paragraph{Efficient Fine-tuning for Memory Encoder.}
To adapt the memory encoder to the agent with minimal cost, we fine-tune it using LoRA~\cite{hu2021lora} (rank 16), updating only 1.2\% of parameters. For data efficiency, we train on 1,500 high-quality trajectories drawn from open-source and synthetic sources. Each step in a trajectory forms a training instance augmented with its top-3 retrieved memories. This parameter- and data-efficient setup completes within 20 hours using NVIDIA H100. More details about training data and settings can be found in Appendix~\ref{append:training_setting}.

\begin{figure}[t]
    \centering
    \includegraphics[width=0.95\linewidth]{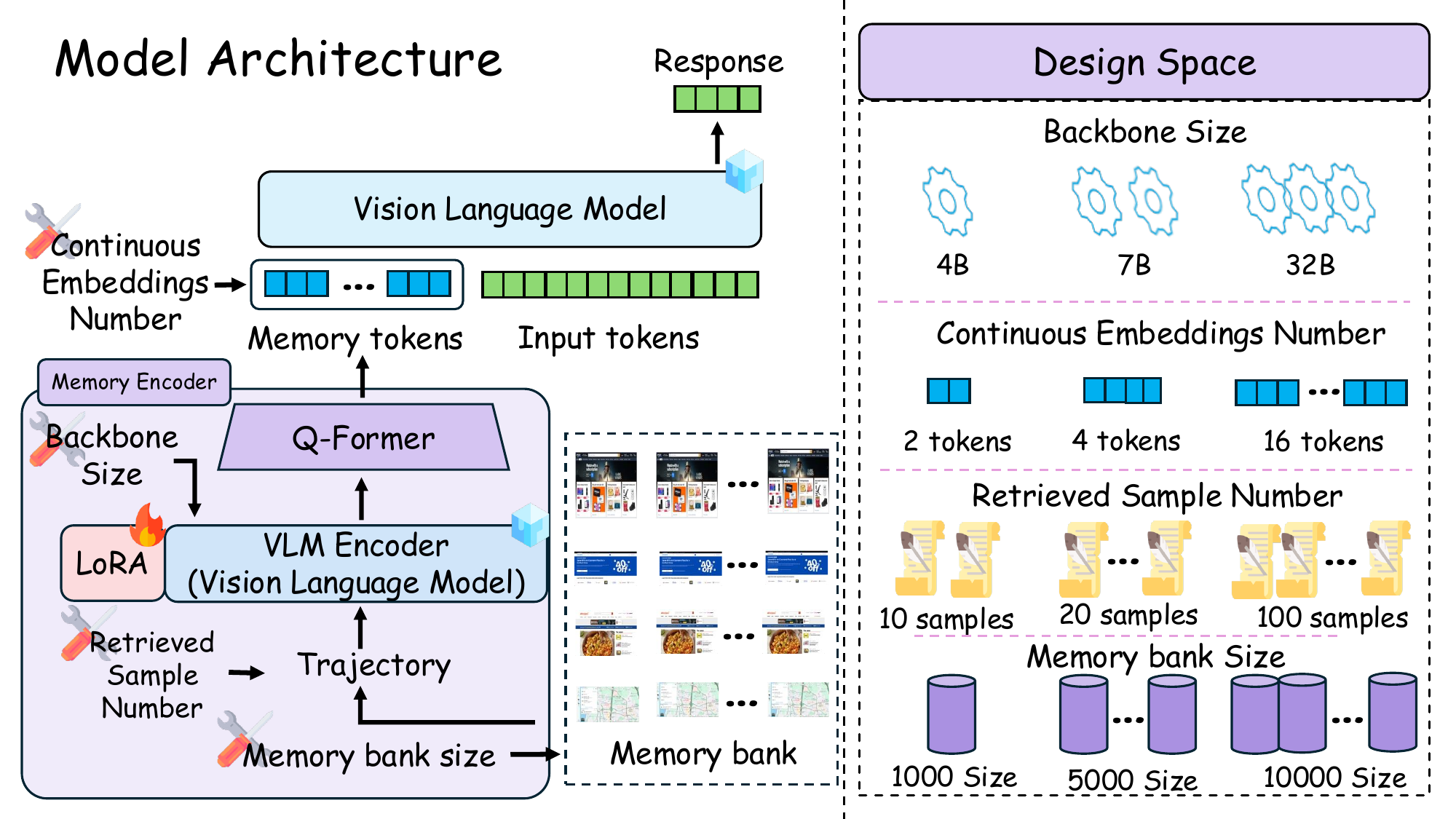}
    \caption{\textbf{Model architecture and explored design space.} A Memory Encoder built from a VLM and Q-Former produces continuous memory embeddings, injected into a frozen VLM at inference. The design space spans backbone size, embedding count, retrieval depth, and memory bank size.}
    \label{fig:architecture}
\end{figure}

\section{Design Space for Scaling Continuous Memory}
\label{sec:scaling_laws}




We systematically study four key design dimensions along which continuous memory can be scaled up for CUAs: the number of retrieved in-context samples $K$, the memory bank size $M$, the backbone model scale $S$, and the number of continuous embeddings per trajectory $N$. For each dimension, we vary across multiple values while holding the others fixed, and examine how CUA success rate responds across different domains, to reveal which dimensions yield clear scale-up trends and which saturate. 
To quantitatively characterize the scaling behavior of continuous memory, we follow established empirical trends in large language modeling and adopt a log-linear framework to describe the relationship between agent performance and our scaling variables. Specifically, we model the Success Rate ($\mathrm{SR}$) as a function of the above variables using the following formulation:

\begin{equation}
\mathrm{SR}(x) = a + b \log x,
\end{equation}

where $b$ denotes the scaling coefficient. We estimate the parameters $\langle a, b \rangle$ using ordinary least squares (OLS) for each dimension independently. 

\subsection{Scaling In-Context Memory Samples}

Building upon our finding that agent trajectories can be represented by highly compressed continuous embeddings, we investigate whether it can be leveraged to include more experience for reference. We evaluate the scaling behavior by varying the number of retrieved memory samples from 3 to 50 across three domains. To isolate the scaling contribution of the memory representation, we compare continuous memory against Reasoning Bank~\cite{ouyang2025reasoningbankscalingagentselfevolving}, a text-memory baseline using the same pool.

\begin{figure}
    \centering
    \includegraphics[width=\linewidth]{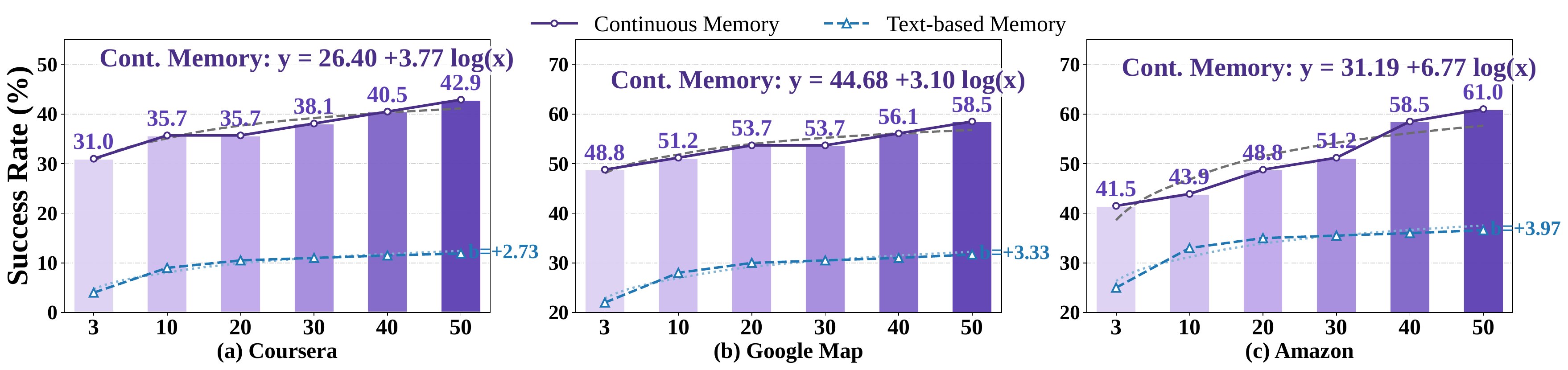}
    \caption{Success rate versus the number of retrieved memory samples with Qwen3-VL-8B. Continuous memory scales more steeply than the text-memory baseline Reasoning Bank.}

    \label{fig:incontext-memory-samples}
\end{figure}

As shown in Figure~\ref{fig:incontext-memory-samples}, continuous memory consistently outperforms Reasoning Bank.
The advantage stems from per-sample context cost: a continuous memory entry occupies only $N=8$ tokens, so the agent can extend to $K=50$ with most of its context window free, whereas text retrievals occupy hundreds of tokens each and rapidly exhaust the context budget at large $K$. On Coursera, the very low baseline means even basic text retrievals supply enough procedural signal to drive comparable per-sample gains, but continuous memory still extracts this signal effectively. 
Even at $K=3$, where text retrievals fit within the context window, continuous memory outperforms Reasoning Bank, indicating that the advantage stems from representation quality, not only context budget.

\subsection{Scaling the Memory Bank Size}

\begin{figure}
    \centering
    \includegraphics[width=\linewidth]{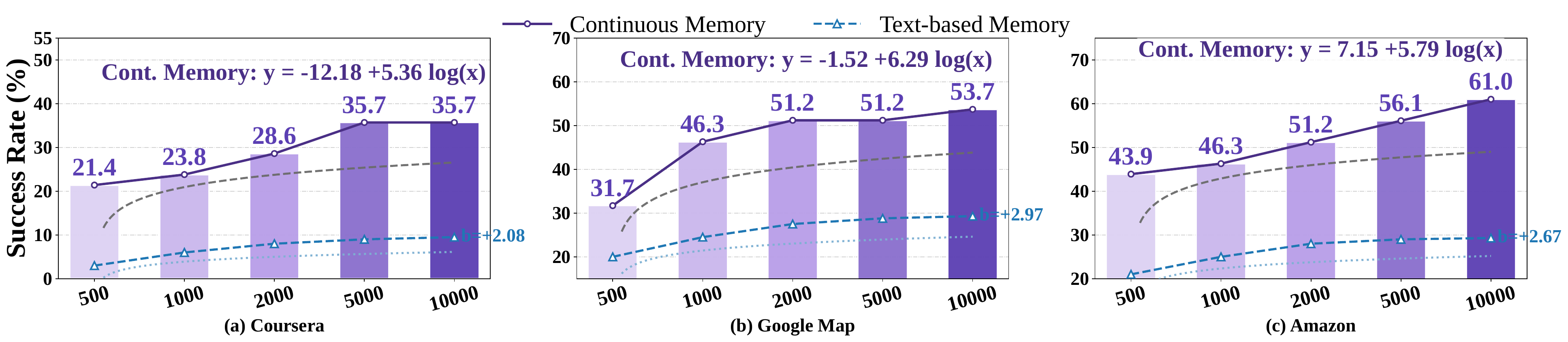}
    \caption{Success rate as the memory bank grows from 500 to 10k across three domains, on Qwen2.5-VL-7B. Continuous memory scales more steeply than the text-memory baseline Reasoning Bank.}

    \label{fig:memory-pool-size}
\end{figure}

In this experiment, we investigate the impact of the memory bank size on agent performance. We scale the memory bank from 500 to 10,000 trajectories, retrieving the most relevant 10 examples for each task. To isolate the scaling contribution of the memory representation itself, we compare continuous memory against Reasoning Bank retrieving from the same pool. As shown in Figure~\ref{fig:memory-pool-size}, both methods improve with a larger bank, but the continuous memory scales more steeply across all three domains: on Google Map, success rises from $31.7\%$ at a 500-item bank to $53.7\%$ at 10k, more than doubling the gain Reasoning Bank achieves over the same sweep.

The mechanism is that each continuous memory entry compactly distills a latent strategy the agent can directly leverage, so a larger bank yields more relevant retrievals, translating into proportionally larger gains. By contrast, text retrievals hand the agent thousands of verbose tokens from which strategy must still be extracted, so the marginal value of additional on-topic retrievals saturates faster.

\subsection{Scaling Backbone Model Size}
\label{sec:scal_backbone}

\begin{wrapfigure}{r}{0.4\textwidth}
    \vspace{-20pt}
    \centering
    \includegraphics[width=\linewidth]{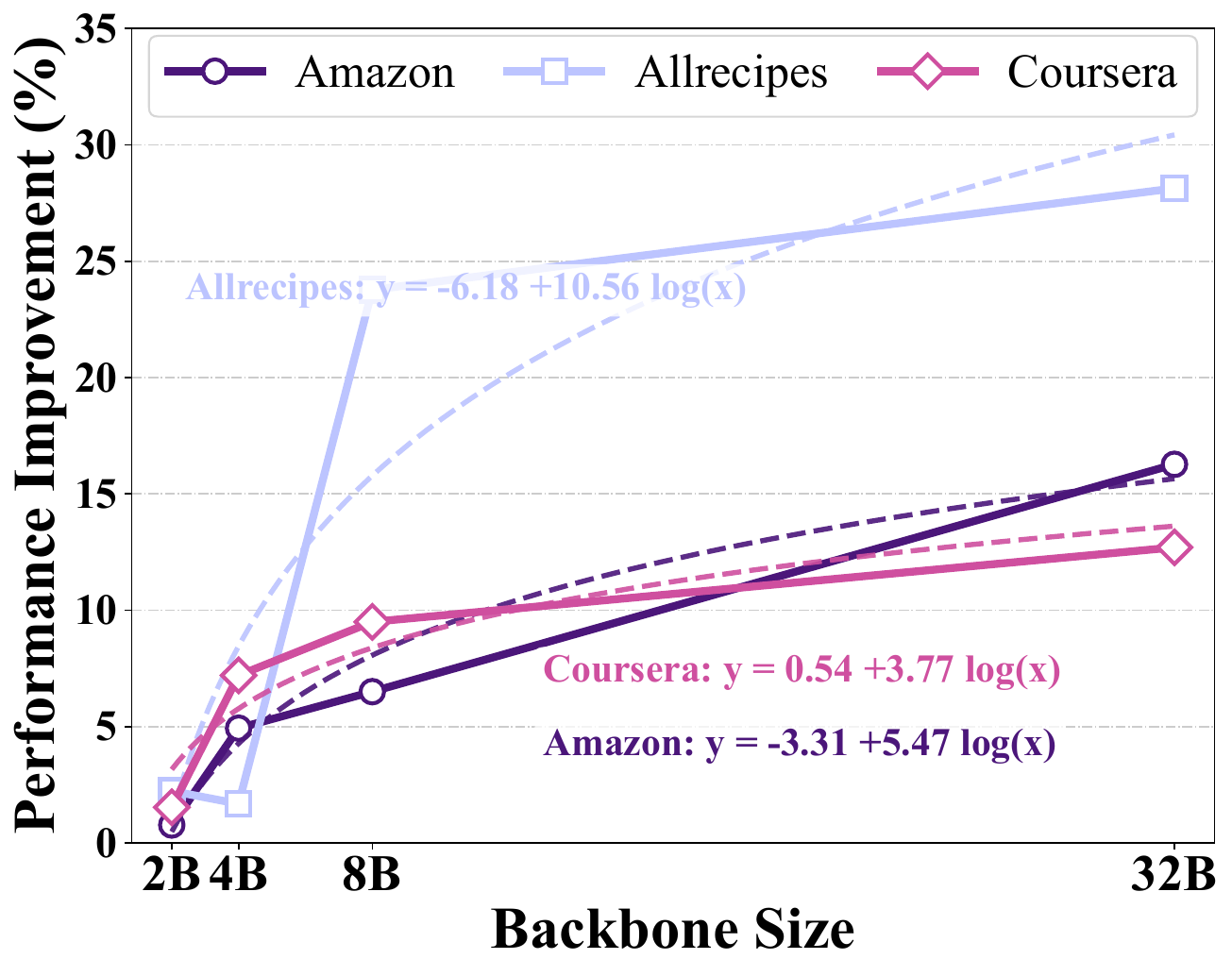}
    \caption{Performance improvement with continuous memory as the backbone model scales.}
    \label{fig:backbone-scale}
    \vspace{-10pt}
\end{wrapfigure}



To evaluate the robustness and scalability of the continuous memory approach, we investigate how the performance gains vary with the capacity of the underlying VLM. In this experiment, we fix the memory retrieval number as 10 and memory bank size as 10k and scale the backbone model size across four configurations: 2B, 4B, 8B, and 32B parameters. 
We measure the absolute performance improvement brought by continuous memory on average over the baseline without memory across three domains.

As Figure~\ref{fig:backbone-scale} shows, the absolute improvement from continuous memory grows with backbone scale: smaller models gain only modestly, while the 32B model reaches nearly $28\%$ improvement on Allrecipes and over $15\%$ on Amazon. The trend reflects how effectively each backbone integrates retrieved cues. Smaller models still make base-level errors in grounding and interaction, so even sound procedural cues retrieved from memory are diluted by execution failures; larger models commit fewer such errors, leaving more headroom for retrieved strategies to translate into success.

\subsection{The Impact of Continuous Embedding Number}

\begin{figure}
    \centering
    \includegraphics[width=\linewidth]{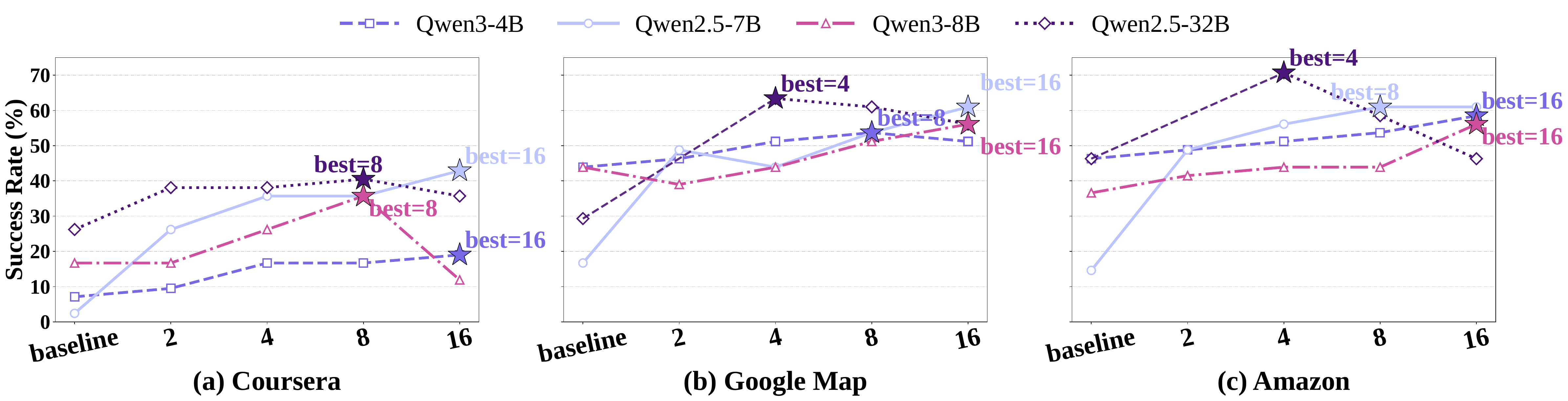}
    \caption{Success rate curve w.r.t. the continuous embedding number across three domains and multiple backbone VLMs. Stars indicate the best-performing results in each setting.}
    \label{fig:compression-token-number}
\end{figure}



We evaluate the impact of continuous embedding number $N \in \{2, 4, 8, 16\}$ across three distinct domains using a suite of VLMs ranging in scale: Qwen3-VL-4B~\cite{yang2025qwen3}, Qwen2.5-VL-7B~\cite{yang2024qwen2.5}, Qwen3-VL-8B~\cite{yang2025qwen3}, and Qwen2.5-VL-32B~\cite{yang2024qwen2.5}. As shown in Figure~\ref{fig:compression-token-number}, these configurations are compared against a no-memory baseline to identify the optimal token budget for representing agent trajectories.

Unlike increases in $K$ and $M$, which yield non-decreasing success rates within the evaluated ranges, $N$ has an interior optimum that shifts with backbone scale. All settings outperform the no-memory baseline, but peak performance is reached at a small budget and degrades as $N$ grows further. The optimal $N$ also decreases with backbone scale: Qwen2.5-VL-32B peaks at $N=4$, while smaller backbones reach their peak at $N=8$ or higher. This pattern reflects a tradeoff between the information capacity of a trajectory encoding and the per-token signal density the agent must attend over. Stronger backbones extract more from each embedding and reach their information ceiling at lower $N$; weaker backbones benefit from more embeddings up to a point, after which redundancy crowds out useful signal.

\section{Autonomous Data Scaling}

The empirical trends in Section~\ref{sec:scaling_laws} show that increasing the in-context retrieval count $K$ and the memory bank size $M$ improves agent success rates within the evaluated ranges. Realizing this benefit at scale requires a large, diverse pool of high-quality trajectories. We therefore build an autonomous data flywheel that grows the memory bank without any human intervention.
For a start, let $\mathcal{Q}$ be the {task/query pool}, $\mathcal{E}$ the {environment pool}, and $\mathcal{T}$ the {trajectory pool}. We initialize $\mathcal{Q}_0$ using a seed query set, \ie Mind2Web training set~\cite{mind2web, zheng2024seeact_multimodal_mind2web}, and set $\mathcal{E}_0$ and $\mathcal{T}_0$ as empty set $\varnothing$.
At each iteration, we update $(\mathcal{Q}_t,\mathcal{E}_t,\mathcal{T}_t)\to(\mathcal{Q}_{t+1},\mathcal{E}_{t+1},\mathcal{T}_{t+1})$ through four phases that continuously discover new environments, synthesize tasks, roll out trajectories, and conduct quality evaluation.

\begin{figure}[ht]
\vspace{-5pt}
    \centering
\includegraphics[width=\linewidth]{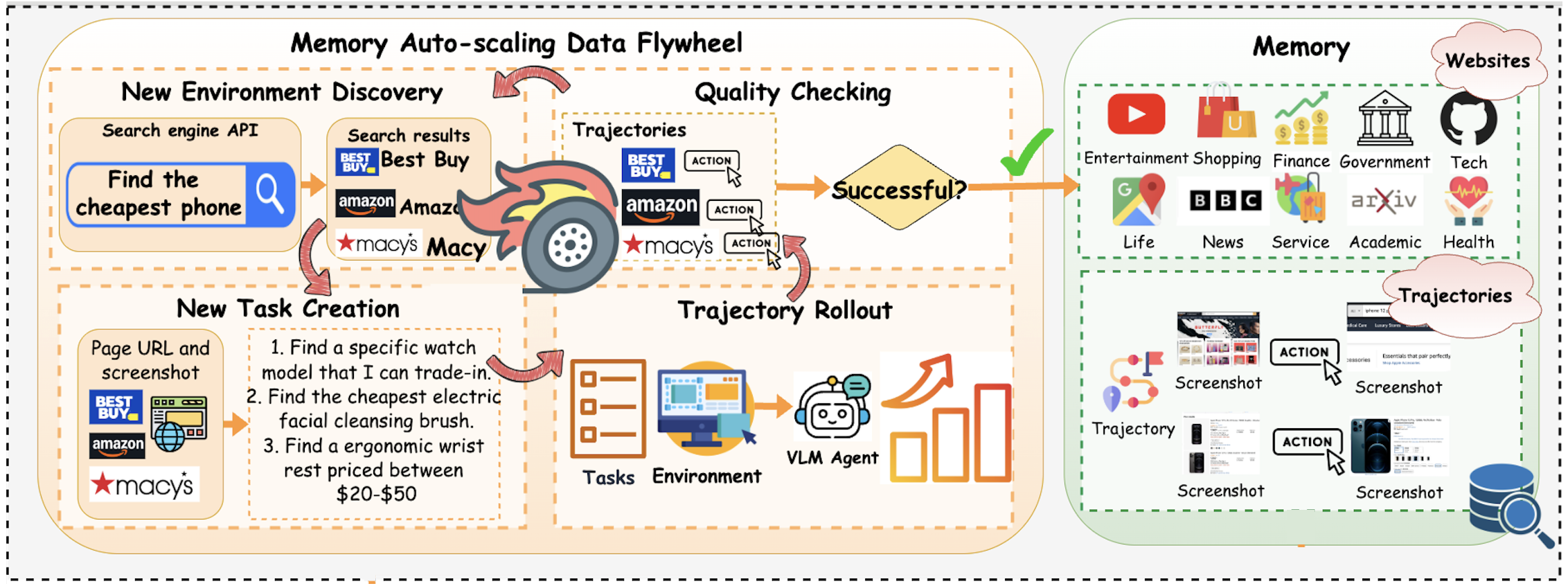}
\caption{Overview of memory auto-scaling data flywheel. We devise (1) a four-phase
memory collection pipeline; (2) a memory bank covering diverse environments and
trajectories.}
    \label{fig:data_flywheel}
\vspace{-10pt}
\end{figure}

\paragraph{Phase-1: New Environment Discovery.}
Inspired by humans who often seek new apps and websites through online search, we utilize an online search engine to discover diverse new environments.
We first sample a few queries from $\mathcal{Q}_t$, and use SerpAPI to search for the top 20 related websites.
Then, we test the stability and accessibility of the websites by using the agent to run test tasks and checking trajectories with VLM, and filter out the low-quality ones.
Next, we deduplicate the websites with $\mathcal{E}_t$, and obtain the newly discovered environment set $\mathcal{E}^{*}$.

\paragraph{Phase-2: New Task Creation.}
For each new environment $e\in {\mathcal{E}}^{*}$, we use a VLM to automatically synthesize new task queries for it. Given the environment screenshot, the VLM first produces a detailed text description of the site. Using this description, along with the screenshot, the VLM then generates a set of candidate task queries $\mathcal{Q}^{*}_e$ that are potentially solvable within the environment.

\paragraph{Phase-3: Trajectory Rollout.}
Next, we employ an agent model (\ie Qwen2.5-VL-32B~\cite{yang2024qwen2.5}) to perform autonomous rollouts.
Given each query $q \in {\mathcal{Q}}^{*}_e$, we follow the setup described in Section~\ref{sec:prelim} to make the agent model interact with environment $e$.
We collect all the actions and observations during the procedure to compose the trajectory $\tau^{*}_{q}=(o_t,a_t)_{t=1}^T$.

\paragraph{Phase-4: Quality Checking.}
Upon completion of each task, we evaluate the resulting trajectory using a dedicated judging model. The evaluation process involves feeding the task query $q$ and the full trajectory $\tau^{*}_{q}$ into the judge VLM, which determines whether the task was successfully completed.
For high-quality evaluation, we use SEAgent-1.0-7B~\cite{SEAgent}, a fine-tuned model specifically trained for CUA assessment. We also use a voting mechanism to ensure robustness.
Finally, we collect all positive trajectories, and their tasks and environments, to update the trajectory pool $\mathcal{T}_t$, task pool $\mathcal{Q}_t$, and environment pool $\mathcal{E}_t$. More details can be found in Appendix~\ref{append:data_quality_control}.

\begin{wraptable}[20]{r}{0.55\textwidth}
\vspace{-15pt}
\caption{Comparison across computer-use data resources. \# Env. and \# Samples denote the number of environments and collected task/trajectory samples. \textbf{Step} denotes whether fine-grained step-wise action trajectories are available, and \textbf{Auto} denotes whether the data collection/annotation process is fully automatic.}
\centering
\footnotesize
\begin{tabular}{l c c c c}
\toprule
\textbf{Dataset} & \textbf{\# Env.} & \textbf{\# Samples} & \textbf{Step} & \textbf{Auto} \\
\midrule
WebArena~\cite{zhou2023webarena}         & 4     & 812            & \cmark & \xmark \\
VisualWebArena~\cite{visualwebarena}   & 3     & 910            & \cmark & \xmark \\
WebShop~\cite{WebShop}          & 1     & 12,087         & \xmark & \xmark \\
OmniACT~\cite{Omniact}          & 30    & 9,802          & \xmark & \xmark \\
OSWorld~\cite{osworld}          & 10    & 369            & \xmark & \xmark \\
ScreenAgent~\cite{niu2024screenagent}      & 39    & 723            & \xmark & \xmark \\
Mind2Web~\cite{mind2web2}         & 137   & 2,350          & \cmark & \xmark \\
WebLINX~\cite{weblinx}          & 150+  & 2,337 & \cmark & \xmark \\
AndroidControl~\cite{androidcontrol}   & 833   & 15,283         & \cmark & \xmark \\
GUIOdyssey~\cite{guiodyssey}       & 201 & 7,735 & \cmark & \xmark \\
InSTA~\cite{trabucco2025instainternetscaletrainingagents}            & 150,000 & 150,000      & \cmark & \cmark \\
\midrule
\textbf{Ours}    & \textbf{35,000+} & \textbf{188,451} & \cmark & \cmark \\
\bottomrule
\end{tabular}
\label{tab:dataset-comparison}
\end{wraptable}

\paragraph{Outcome.}
This \emph{discover} $\rightarrow$ \emph{generate} $\rightarrow$ \emph{rollout} $\rightarrow$ \emph{verify} closed loop provides a flexible auto-scaling strategy to continuously expand $\mathcal{Q}$, $\mathcal{E}$, and $\mathcal{T}$.
In practice, by setting simple diversity and quality filters with rule-based and LLM-as-Judge methods~\cite{gu2024llmjudge}, we can prevent redundancy and ensure robustness.
In experiments, we spent \$1,972 USD to collect 188,451 trajectories spanning over 35,000 environments.
As shown in Table~\ref{tab:dataset-comparison}, our trajectory dataset is among the largest fully automatic, step-wise computer-use data resources, and is constructed as a scalable memory bank for continuous-memory augmentation.
Compared with human-collected datasets, our data flywheel provides substantially broader environment coverage and lower annotation cost.
Compared with automatic web-only pipelines such as InSTA, our pipeline is designed to collect high-quality multimodal trajectories for memory construction, retrieval, and continuous embedding compression.
More details are in Appendix~\ref{sec:task generation pipeline}.

\section{Experiments}
\subsection{Experimental Setup}
\label{sec:env setup}

\paragraph{Memory Settings.}
\label{sec:exp_memory_setting}
We construct the memory bank from successful trajectories using the data flywheel. For our experiments in the web-browser setting, we utilize a randomly selected subset of 10,000 trajectories from this bank. For the experiment in the OS setting, we use GUIAct~\cite{chen2025guiact} dataset as our memory bank. For each task, we retrieve $k=10$ memory examples and compress each trajectory into an 8-token embedding.
Note that while we do not perform online memory eviction during inference, we apply offline pruning and deduplication during memory construction to ensure high quality: (1) Only successful trajectories are retained. (2) Rule-based heuristics remove redundant or erroneous steps (\eg repeated clicks, ineffective scrolling). (3) Domain-aware filtering is applied during retrieval to ensure relevance.
These combined steps help maintain memory quality and prevent low-value or noisy entries from polluting the memory bank.
Further details regarding memory bank construction, quality control, and selection criteria are provided in Appendix~\ref{append:memory_construct}.

\paragraph{Evaluation Tasks.}
We evaluate our approach across both web-browser and operating system (OS) scenarios. Our evaluation suite includes three multimodal web-browser benchmarks and one OS benchmark. For the web-browser tasks, we use WebVoyager~\cite{he2024webvoyager}, Mind2Web~\cite{mind2web}, and MMInA~\cite{zhang2024mmina}. For OS-level evaluation, we employ the OSWorld~\cite{osworld} benchmark. We report task accuracy, evaluated according to the standard protocols of each benchmark, as our primary metric. A detailed description of these benchmarks can be found in Appendix~\ref{append:evaluation_benchmark}.


\paragraph{Baseline Methods.}
We compare our method with closed-source models, task-specialized open-source CUAs, and open-source models augmented with text-based memory. Details and fairness considerations are provided in Appendix~\ref{append:baseline_methods}. All results are from our reproduction. The environment setup, task sampling, and evaluation protocols are consistent for fair comparison across models.


\subsection{Main Results}

\begin{table*}[t]
\centering
\scriptsize
\setlength{\tabcolsep}{2pt}
\renewcommand{\arraystretch}{1.05}
\setlength{\aboverulesep}{0.25ex}
\setlength{\belowrulesep}{0.25ex}

\caption{Task success rates (\%) across WebVoyager, Mind2Web, and MMInA. Bolded entries indicate best-performing variant across \{Baseline, +Reasoning Bank, +AWM, +Ours\} in each domain.}
\resizebox{\textwidth}{!}{%
\begin{tabular}{@{} l l *{10}{c} @{}}
\toprule
\multicolumn{2}{c}{}
& \multicolumn{4}{c}{\textbf{\shortstack{WebVoyager}}}
& \multicolumn{5}{c}{\textbf{\shortstack{Mind2Web \& MMInA}}} \\
\cmidrule(lr){3-6} \cmidrule(lr){7-11}
\textbf{Backbone} & \textbf{\shortstack{Model/Method}}
& \textbf{Amz} & \textbf{Cour} & \textbf{Recp} & \textbf{Map}
& \textbf{Info} & \textbf{Svc} & \textbf{Ent} & \textbf{Trav} & \textbf{Wiki}
& \textbf{Avg.} \\
\midrule

\multirow{3}{*}{Closed-Source}
& GPT-4o            & 24.4\% & 7.1\%  & 13.3\% & 36.6\% & 7.8\%  & 14.1\% & 3.4\%  & 19.4\% & 51.0\% & 19.7\% \\
& Gemini-2.5-Pro & 41.7\% & 42.9\% & 20.0\% & 53.7\% & 16.7\% & 22.4\% & 0.0\%  & 19.4\% & 50.0\% & 29.6\% \\
& Claude-4          & 63.4\% & 28.6\% & 33.3\% & 70.0\% & 25.6\% & 40.0\% & 6.9\%  & 9.7\%  & 52.0\% & 36.6\% \\
\midrule

\multirow{3}{*}{Open-Source}
& UI-TARS-1.5-7B    & 31.7\% & 16.7\% & 20.0\% & 31.7\% & 6.4\%  & 4.7\%  & 6.9\%  & 0.0\%  & 36.0\% & 17.1\% \\
& CogAgent          & 12.2\% & 9.5\%  & 26.7\% & 9.8\%  & 24.4\% & 8.2\%  & 13.8\% & 16.1\% & 21.0\% & 15.7\% \\
& Websight          & 24.4\% & 4.8\%  & 13.3\% & 29.3\% & 10.3\% & 3.5\%  & 3.4\%  & 0.0\%  & 12.0\% & 11.2\% \\
\midrule

\multirow{4}{*}{Qwen3-VL-2B}
& Baseline          & 36.6\% & 16.7\% & 17.8\% & 36.6\% & 12.8\% & 8.2\% & 0.0\% & 16.1\% & 39.0\%  & 20.4\% \\
& + Reasoning Bank  & \textbf{43.9\%} & 21.4\% & 11.1\% & 41.5\% & 11.5\% & 11.8\% & 0.0\% & \textbf{19.4\%} & \textbf{45.0\%} & 22.8\% \\
& + AWM             & 29.3\% & \textbf{23.8\%} & 13.3\% & 39.0\% & 9.0\% & \textbf{21.2\%} & \textbf{3.4\%} & 16.1\% & 41.0\% & 21.8\% \\
\rowcolor{cyan!8}
& + \textbf{Ours}  & 39.0\% & 21.4\% & \textbf{28.9\%} & \textbf{48.8\%} & \textbf{16.7\%} & 17.6\% & \textbf{6.9\%} & 12.9\% & 44.0\% & \textbf{26.2\%} \\
\midrule

\multirow{4}{*}{Qwen3-VL-4B}
& Baseline          & 46.3\% & 7.1\%  & 20.0\% & 43.9\% & 7.7\% & 16.5\% & \textbf{10.3\%} & 12.9\% & 41.0\% & 22.9\% \\
& + Reasoning Bank  & 51.2\% & 11.9\% & 15.6\% & 51.2\% & 11.5\% & \textbf{24.7\%} & 6.9\% & 3.2\% & 45.0\% & 24.6\% \\
& + AWM             & 48.8\% & 11.9\% & 17.8\% & 51.2\% & 14.1\% & 20.0\% &  3.4\% & 6.5\% & \textbf{51.0\%} & 25.0\% \\
\rowcolor{cyan!8}
& + \textbf{Ours}  & \textbf{53.7\%} & \textbf{16.7\%} & \textbf{22.2\%} & \textbf{53.7\%} & \textbf{16.7\%} & 23.5\% & 6.9\% & \textbf{12.9\%} & 42.0\% & \textbf{27.6\%} \\
\midrule

\multirow{4}{*}{Qwen2.5-VL-7B}
& Baseline          & 14.6\% & 2.4\%  & 15.9\% & 16.7\% & 9.0\%  & 11.8\% & 0.0\%  & 4.4\%  & 38.0\% & 12.5\% \\
& + Reasoning Bank  & 29.3\% & 9.5\%  & 6.7\%  & 29.3\% & 9.0\%  & 20.0\% & 3.4\%  & 6.5\%  & 44.0\% & 17.5\% \\
& + AWM             & 17.1\% & 4.8\%  & 11.1\% & 29.3\% & 7.7\%  & 10.6\% & 0.0\%  & 6.5\%  & 31.0\% & 13.1\% \\
\rowcolor{cyan!8}
& + \textbf{Ours}  & \textbf{53.6\%} & \textbf{35.7\%} & \textbf{31.1\%}  & \textbf{53.7\%} & \textbf{21.8\%} & \textbf{27.1\%} & \textbf{6.9\%} & \textbf{9.7\%} & \textbf{47.0\%} & \textbf{31.9\%} \\
\midrule

\multirow{4}{*}{Qwen3-VL-8B}
& Baseline          & 36.6\% & 16.7\% & 13.3\% & 43.9\% & 9.0\%  & 20.0\% & 10.3\% & 9.7\%  & 38.0\% & 21.9\% \\
& + Reasoning Bank  & 36.6\% & 11.9\% & 17.8\% & 31.7\% & 10.3\% & \textbf{24.7\%} & 17.2\% & 6.5\% & 48.0\%& 22.7\% \\
& + AWM             & 39.0\% & 19.0\% & 8.9\%  & 31.7\% & 12.5\% & 16.5\% & 6.9\%  & 16.1\% & 48.0\% & 22.1\% \\
\rowcolor{cyan!8}
& + \textbf{Ours}  & \textbf{43.9\%} & \textbf{35.7\%} & \textbf{37.8\%} & \textbf{51.2\%} & \textbf{15.4\%} & 22.4\% & 17.2\% & 16.1\% & \textbf{52.0\%} & \textbf{32.4\%} \\
\midrule

\multirow{4}{*}{Qwen2.5-VL-32B}
& Baseline          & 46.3\% & \textbf{26.2\%} & 6.7\%  & 29.3\% & 14.1\% & 20.0\% & 6.9\%  & 9.7\%  & 43.0\% & 22.5\% \\
& + Reasoning Bank  & 58.5\% & 21.4\% & 24.4\% & 56.1\% & \textbf{15.4\%} & 24.7\% & 3.4\% & 12.9\% & 39.0\% & 28.4\% \\
& + AWM             & 48.8\% & 26.2\% & 28.9\% & 51.2\% & 12.8\% & 18.8\% & 6.9\% & 6.5\% & 45.0\% & 27.2\% \\
\rowcolor{cyan!8}
& + \textbf{Ours}  & 58.5\% & \textbf{40.5\%} & \textbf{35.6\%} & \textbf{61.0\%} & 12.8\% & 24.7\% & \textbf{10.3\%} & 12.9\% & \textbf{49.0\%} & \textbf{33.9\%} \\
\bottomrule
\end{tabular}%
}
\label{tab:memory-model-comparison}
\end{table*}

\begin{table*}[t]
\centering
\normalsize
\renewcommand{\arraystretch}{1.20}
\setlength{\tabcolsep}{2.2pt}
\setlength{\aboverulesep}{0.35ex}
\setlength{\belowrulesep}{0.35ex}
\caption{OSWorld success rates (\%) across application suites.}
\resizebox{\textwidth}{!}{%
\begin{tabular}{@{}l c c c c c c c c c c c c c c@{}}
\toprule
\multirow{2}{*}{\textbf{Method}}
& \multicolumn{4}{c}{\textbf{Office}}
& \multicolumn{4}{c}{\textbf{Daily}}
& \multicolumn{3}{c}{\textbf{Prof.}}
& \multicolumn{2}{c}{\textbf{Others}}
& \multirow{2}{*}{\textbf{Overall}} \\
\cmidrule(lr){2-5}\cmidrule(lr){6-9}\cmidrule(lr){10-12}\cmidrule(lr){13-14}
& \textbf{Writer} & \textbf{Calc} & \textbf{Impress} & \textbf{Avg.}
& \textbf{Thunderbird} & \textbf{VLC} & \textbf{Chrome} & \textbf{Avg.}
& \textbf{GIMP} & \textbf{VS Code} & \textbf{Avg.} & \textbf{M-Apps} & \textbf{OS}
&   \\
\midrule

\textbf{Qwen2.5-VL-7B}               & 4.3\%  & 2.1\% & 2.1\% & 2.6\% & 0.0\%  & 5.9\% & 11.1\% & 7.8\% & 0.0\%  & 0.0\% & 0.0\% & \textbf{2.0\%} & 12.5\% & 3.9\% \\
+ Reasoning Bank       & 8.7\%  & 2.1\% & 2.1\% & 3.4\% & 13.3\% & 0.0\% & 6.5\%  & 6.4\% & 11.5\% & 4.6\% & 8.3\% & 0.9\% & 12.5\% & 4.7\% \\
\rowcolor{orange!15}
+ \textbf{Ours}       & \textbf{13.0\%} & \textbf{6.4\%} & \textbf{8.5\%} & \textbf{8.6\%} & \textbf{20.0\%} & 5.9\% & \textbf{19.6\%} & \textbf{16.7\%} & \textbf{15.4\%} & 4.6\% & \textbf{10.4\%} & 0.9\% & 12.5\% & \textbf{8.9\%} \\
\midrule

\textbf{Qwen3-VL-8B}                 & 27.3\% & 4.3\% & \textbf{19.0\%} & \textbf{14.6\%} & 40.0\% & 17.7\% & 28.3\% & 28.2\% & 30.8\% & 18.2\% & 25.0\% & 7.4\% & 29.2\% & 18.1\% \\
+ Reasoning Bank         & 18.2\% & \textbf{8.5\%} & 10.6\% & 11.2\% & \textbf{53.3\%} & 23.5\% & 26.1\% & 30.8\% & 36.0\% & \textbf{27.3\%} & 31.9\% & 6.3\% & 41.7\% & 19.0\% \\
\rowcolor{orange!15}
+ \textbf{Ours} & 27.3\% & 6.5\% & 10.4\% & 12.1\% & 46.7\% & 23.5\% & \textbf{34.8\%} & \textbf{34.6\%} & \textbf{46.2\%} & 18.2\% & \textbf{33.3\%} & \textbf{8.5\%} & 41.7\% & \textbf{20.9\%} \\
\bottomrule
\end{tabular}%
}
\label{tab:osworld_results}
\end{table*}


\paragraph{Web-Browser Environments.}
Open-source backbones achieve only modest performance on these benchmarks, and existing text-based memory provides a limited uplift. Our method improves every tested backbone, with memory-augmented success rates ranging from $26.2\%$ on the 2B backbone to $33.9\%$ on the 32B variant in Table~\ref{tab:memory-model-comparison}. Section~\ref{sec:scal_backbone} examines how the gains vary with backbone scale. Equipped with our method, the 32B open-source backbone achieves the best performance among the open-source models we evaluate, matching or surpassing GPT-4o and Gemini-2.5-Pro and approaching Claude-4 ($33.9\%$ vs.\ $36.6\%$) without any base-model fine-tuning.

The gains are not uniform across domains. They concentrate on procedure-rich tasks where successful trajectories share reusable multi-step search--filter--act patterns, \eg on the 7B backbone, Coursera improves by over 30\%. By contrast, lookup-heavy domains where each task reduces to short retrieval steps benefit much less, \eg Wikipedia rises by under 10 points.
 A small number of cells regress, primarily when the retrieval pool is sparse for the target domain or when short, rule-driven tasks are better served by an explicit textual cue. 



\paragraph{Operating System Environment.}
On the more challenging OSWorld benchmark, our method transfers cleanly to the desktop setting and lifts Qwen2.5-VL-7B from $3.9\%$ to $8.9\%$, more than doubling its overall success rate under the fixed-framework evaluation protocol detailed in Appendix~\ref{append:osworld_protocol}. On the stronger Qwen3-VL-8B backbone, gains concentrate in the Daily and Professional suites, tasks involving genuine multi-step procedural transfer such as media playback, browser automation, and image editing, while the Office suite shows little or no benefit, since these tasks consist of short, rigid menu sequences with limited cross-task transfer. 

Together, these results show where our method adds the most value: tasks with a multi-step procedural structure that compresses and transfers across episodes. Additional analyses on inference latency, noisy memory robustness, and learning from errors are provided in Appendix~\ref{append:further_analysis}.

\section{Related Work}

\paragraph{Computer-Use Agents.}
Recent benchmarks have standardized CUA evaluation across web, desktop, and mobile settings, including Mind2Web~\cite{mind2web}, WebArena~\cite{zhou2023webarena}, VisualWebArena~\cite{visualwebarena}, OSWorld~\cite{osworld, bonatti2024windowsagentarenaevaluating}, and GUIOdyssey~\cite{guiodyssey}. Modern CUAs improve through stronger grounding~\cite{screenspotpro,osatlas}, larger-scale training~\cite{uitars, ye2025mobileagentv3fundamentalagentsgui}, and RL-style fine-tuning~\cite{luo2025guir1, lu2025uir1}. Representative systems include text-centric agents such as WebGPT~\cite{WebGPT} and WebShop~\cite{WebShop}, as well as screenshot-based agents such as SeeAct~\cite{zheng2024seeact_multimodal_mind2web}, WebSight~\cite{bhathal2025websight}, and UI-TARS~\cite{uitars}. Planner Matters!~\cite{wu2026plannermattersefficientunbalanced} studies planner-centric compute allocation and reinforcement learning in a planner--actor--memory-manager framework. StructAgent~\cite{wu2026structagent} organizes long-horizon execution around a unified state and verifier-backed workflow, supporting progress tracking and failure recovery. These approaches address planning and execution structure, complementing our study of scalable continuous-memory representations.

\paragraph{Memory for LLMs and Agents.}
External memory has been widely studied to address finite context windows and weak long-term retention. Retrieval-based methods such as RAG~\cite{RAG2020} and REALM~\cite{realm2020} use non-parametric textual memory, but long concatenated contexts increase inference cost and may distract planning~\cite{RAGsurvey2024}. Recent work explores structured or continuous memories, including compact visual embeddings~\cite{VoCo-LLaMA, he2024malmm}, portable VLM memory~\cite{wu2025}, workflow memory~\cite{wang2024agentworkflowmemory}, procedural memory~\cite{fang2025memp}, and multimodal trajectory retrieval~\cite{zhang2025universalretrievalmultimodaltrajectory}. HyMEM~\cite{zhu2026hybridselfevolvingstructuredmemory} combines symbolic nodes with continuous trajectory embeddings in a self-evolving graph that supports multi-hop retrieval and working-memory updates. Our study focuses on a complementary question: how retrieval depth, bank size, backbone scale, and embedding count affect continuous-memory augmentation for multimodal CUAs.

\paragraph{Scalable Data Generation for Agent Memory.}
LLM-driven synthesis and self-improvement reduce supervision costs, from instruction bootstrapping~\cite{wang-etal-2023-self-instruct, zeng-etal-2024-automatic, xu2024wizardlm} to computer-use data generation and self-evolution~\cite{ZeroGUI, SEAgent}. Related benchmarks use VLMs for trajectory-level evaluation~\cite{mind2web,mind2web2,he2024webvoyager}. RSIAgent~\cite{zhu2026rsiagent} coordinates curriculum, actor, and verifier agents for broad-then-deep exploration and environment-specific memory construction, then reuses the frozen memory for downstream tasks. Our data flywheel follows this experience-acquisition paradigm, but targets autonomous expansion and verification of a large memory bank for studying how continuous memory scales with experience under increasing data budgets, diverse task distributions, and long-horizon interaction settings.

\section{Conclusion}
We presented a memory-augmented framework for CUAs that tackles long-horizon tasks by encoding prior trajectories as compact continuous embeddings and scaling this memory through an autonomous data flywheel. Continuous memory substantially compresses multimodal trajectories while preserving critical visual cues, enabling practical retrieval augmentation at inference without ballooning context length. We observed positive empirical scaling trends in retrieval depth, memory size, and backbone size within the evaluated ranges, while the embedding count has a backbone-dependent optimum. To sustain growth, we devised an auto-scaling data flywheel to collect diverse and high-quality experiences, consisting of environment discovery, task synthesis, trajectory rollout, and quality checking. Using only a search engine, an open-source VLM, and an agent, our data flywheel yielded 188,451 trajectories at low cost. Based on the collected memory, we adopted a lightweight regime (LoRA + Q-Former; 1.2\% parameters on 1.5k samples) to equip open-source models with memory. Experimental results show that the memory-augmented model can attain accuracy competitive with leading closed-source models.

Overall, our results indicate that continuous memory and autonomous experience acquisition form a robust foundation for scalable, generalizable CUAs.
Future work includes adaptive retrieval policies conditioned on uncertainty, tighter integration with RL for credit assignment over memories, extension to mobile platforms, and privacy-aware on-device memorization.

\begin{ack}
Acknowledgments will be added in the camera-ready version.
\end{ack}

\bibliographystyle{unsrtnat}
\bibliography{main}


\clearpage

\appendix

\section{Baseline Methods}
\label{append:baseline_methods}

To assess the effectiveness of our memory-augmented framework, we compare the performance across four settings. \textbf{(i) Closed-source models} include GPT-4o~\cite{openai2024gpt4ocard}, Gemini-2.5-Pro~\cite{comanici2025gemini}, and Claude-4~\cite{anthropic2025claude4}. \textbf{(ii) Open-source models} include UI-TARS-1.5~\cite{uitars}, CogAgent~\cite{hong2023cogagent}, and WebSight~\cite{bhathal2025websight}, which are fine-tuned specifically for computer-use tasks and provide a comparison with task-specific adaptation. \textbf{(iii) Open-source models with text-based memory} include ReasoningBank~\cite{ouyang2025reasoningbankscalingagentselfevolving} and Agent Workflow Memory (AWM)~\cite{wang2024agentworkflowmemory}, which distill experience from past trajectories into tokenized textual prompts. \textbf{(iv) Open-source models with continuous memory} adopt the CoMEM framework~\cite{wu2025}, which compresses and stores multimodal external experience trajectories, including text and screenshots, in a continuous embedding space to evaluate the benefits of rich multimodal memory representations for complex agent tasks. We refer to this configuration evaluated under our scaled bank as Ours.

\paragraph{Fairness for memory modality.}
Although Reasoning Bank and AWM are applicable and evaluated on GUI agents, they store memory as discrete text tokens, verbalizing trajectories into textual descriptions before indexing and retrieval. We compare against these text-only formulations because adapting them to natively encode visual cues would require representational changes that fundamentally converge toward continuous multimodal memory. The comparison therefore characterizes the inherent gap between text-only and continuous-multimodal memory representations, not implementation differences.

\paragraph{Fairness for closed-source comparison.}
All baselines, including closed-source models, run in a unified inference framework with the same prompting template, screenshot-based perception, and action-parsing pipeline; no per-backbone harness or prompt optimization is applied. The closed-source numbers we report therefore measure backbone capability under matched conditions, not the leaderboard maxima reachable with carefully engineered agent frameworks.

\paragraph{OSWorld comparison protocol.}
\label{append:osworld_protocol}
State-of-the-art performance on the OSWorld leaderboard is achieved by closed-source frontier models combined with carefully engineered, OS-specific agent frameworks. In contrast, we use only open-source small-to-medium-sized backbones with a standard inference loop and no framework-level optimization. Therefore, our absolute OSWorld scores are below leaderboard SOTA by design; the goal is to isolate the relative contribution of our method under fixed-framework conditions.

\section{Further Analysis}
\label{append:further_analysis}

\begin{table}[htbp]
\vspace{-20pt}
\centering
\caption{Inference speed and performance comparison across different models. We report the average inference time (in seconds) per trajectory and accuracy (Acc.).}
\resizebox{0.9\textwidth}{!}{%
\begin{tabular}{l c c c c c c c c}
\toprule
\multirow{2}{*}{\textbf{Backbone}} & \multicolumn{2}{c}{\textbf{Amazon}} & \multicolumn{2}{c}{\textbf{Google Map}} & \multicolumn{2}{c}{\textbf{Coursera}} & \multicolumn{2}{c}{\textbf{Allrecipes}} \\
\cmidrule(lr){2-3} \cmidrule(lr){4-5} \cmidrule(lr){6-7} \cmidrule(lr){8-9}
& \textbf{Time (s)} & \textbf{Acc.} & \textbf{Time (s)} & \textbf{Acc.} & \textbf{Time (s)} & \textbf{Acc.} & \textbf{Time (s)} & \textbf{Acc.} \\
\midrule
GPT-4o  & 323.5 & 24.4\% & 248.6 & 36.6\% & 272.2 & 7.1\%  & 295.4 & 13.3\% \\
Claude-4 & 100.7 & 63.4\% & 91.6  & 70.0\% & 161.1 & 28.6\% & 174.1 & 33.3\% \\
Qwen2.5-VL-7B & 163.8 & 14.6\% & 162.8 & 16.7\% & 134.4 & 2.4\%  & 209.1 & 15.9\% \\
~~+Ours    & 154.3 & 53.6\% & 162.6 & 53.7\% & 183.6 & 35.7\% & 355.3 & 31.1\% \\
\bottomrule
\end{tabular}
}
\label{tab:efficiency}
\vspace{-10pt}
\end{table}

\paragraph{Inference Latency Study.}
To assess the impact of memory augmentation on agent efficiency, we measured the average inference time and accuracy per trajectory across four domains: Amazon, Google Map, Coursera, and Allrecipes. The results, presented in Table~\ref{tab:efficiency}, reveal a nuanced relationship between memory augmentation, latency, and performance.

%

Wall-clock latency is essentially unchanged on Amazon and Google Maps, but rises moderately on Coursera and Allrecipes, where the memory-augmented agent traverses additional steps to reach success states that the baseline misses. Even on these higher-latency trajectories, the success-rate gain outweighs the time overhead: on Coursera, for instance, the agent succeeds nearly fifteen times more often, from $2.4\%$ to $35.7\%$ with less than 50 seconds of additional time. The moderate latency increase is therefore a worthwhile trade for the substantial gain in success rate.


\begin{table}[h!]
\vspace{-10pt}
\centering
\caption{Stress Test of Ours under Noisy Memory}
\label{tab:noisy_memory_stress_test}
\resizebox{0.9\textwidth}{!}{%
\begin{tabular}{lcccccc}
\toprule
\textbf{Model} & \textbf{Amazon} & \textbf{Map} & \textbf{Allrecipes} & \textbf{Info} & \textbf{Service} & \textbf{Wikipedia} \\
\midrule
Qwen2.5-VL-7B & 14.6\% & 16.7\% & 15.9\% & 9.0\% & 11.8\% & 38.0\% \\
+ Reasoning Bank & 29.3\% & 29.3\% & 6.7\% & 9.0\% & 20.0\% & 44.0\% \\
+ Ours & 53.6\% & 51.2\% & 37.8\% & 15.4\% & 22.4\% & 52.0\% \\
+ Noisy Memory + Ours & 48.8\% & 51.2\% & 33.3\% & 12.8\% & 17.6\% & 51.0\%\\
\bottomrule
\end{tabular}
}
\end{table}

\vspace{-10pt}
%

\paragraph{Robustness to Noisy Memory.}
To stress-test our memory mechanism we randomly replaced 2 of 10 retrieved samples with low-similarity ones. As shown in Table~\ref{tab:noisy_memory_stress_test}, the noisy variant retains most of the clean-memory advantage and remains well above the no-memory baseline; even under this perturbation, it still outperforms text-based Reasoning Bank by over $12$ percentage points. This robustness comes from representation rather than retrieval relevance: every entry distills a latent strategy from a successful trajectory, so even off-topic retrievals carry transferable interaction patterns.


\begin{table}[h!]
\vspace{-10pt}
\centering
\caption{Performance Comparison with Learn-from-Error (Success Rate \%)}
\label{tab:learn-from-error}
\resizebox{\textwidth}{!}{%
\begin{tabular}{llcccccc}
\toprule
\textbf{Backbone} & \textbf{Setting} & \textbf{Amazon} & \textbf{Coursera} & \textbf{Recipes} & \textbf{Google Map} & \textbf{Info} & \textbf{Services} \\
\midrule
\multirow{4}{*}{Qwen2.5-VL-7B} & Baseline & 14.6\% & 2.4\% & 15.9\% & 16.7\% & 9.0\% & 11.8\% \\
& Reasoning Bank & 29.3\% & 9.5\% & 6.7\% & 29.3\% & 7.7\% & 10.6\% \\
 & Ours & 53.6\% & 35.7\% & \textbf{31.1\%} & 53.7\% & 16.7\% & 23.5\% \\
 & Ours+Learn from Error & \textbf{63.4\%} & \textbf{54.8\%} & 20.0\% & 53.7\% & \textbf{17.9\%} & 23.5\% \\
\midrule
\multirow{4}{*}{Qwen3-VL-8B} & Baseline & 36.6\% & 16.7\% & 13.3\% & 43.9\% & 9.0\% & 20.0\% \\
& Reasoning Bank & 36.6\% & 11.9\% & 17.8\% & 31.7\% & 10.3\% & \textbf{24.7\%} \\
 & Ours& 43.9\% & \textbf{35.7\%} & \textbf{37.8\%} & 51.2\% & 15.4\% & 22.4\% \\
 & Ours+Learn from Error & \textbf{56.1\%} & 33.3\% & 35.6\% & 51.2\% & \textbf{24.4\%} & 23.5\% \\
\bottomrule
\end{tabular}
}
\end{table}


\paragraph{Integration with Experience from Error.}
While our main setting focuses on successful trajectories for high-quality experience, we additionally validate that our method can incorporate negative signal naturally. We introduce an extension, Ours + Learn-from-Error, which distills failure takeaways from negative trajectories. At inference time, the agent is injected with $k_P=10$ successful continuous embeddings and prompted with failure takeaways distilled from $k_N=10$ retrieved negative trajectories. Table~\ref{tab:learn-from-error} compares this extension against the baseline, the original setting, and the pure-text Reasoning Bank~\cite{ouyang2025reasoningbankscalingagentselfevolving}.

The effect is domain-dependent rather than uniform. On the 7B backbone, Coursera gains another $+19$pp beyond the strong original result, while Recipes regresses by roughly $11$pp. The pattern reflects a saturation effect: where the baseline still has diverse failure modes to learn from, failure takeaways add substantial value; where the dominant procedure is already captured by Ours with successful trajectories alone, additional failure cues function as context-dependent noise. Even so, the extended variant outperforms text-only Reasoning Bank in 11 of 12 settings, supporting the value of pairing successful continuous experience with explicit failure cues.

\section{Evaluation Benchmark}
\label{append:evaluation_benchmark}

Experiments are conducted on four challenging multimodal agent benchmarks: WebVoyager~\cite{he2024webvoyager}, Multimodal-Mind2Web~\cite{mind2web, zheng2024seeact_multimodal_mind2web}, MMInA~\cite{zhang2024mmina}, and OSWorld~\cite{osworld}, covering diverse real-world computer-use scenarios.

\begin{itemize}
    \item[(i)] \textbf{WebVoyager}~\cite{he2024webvoyager} contains real-world tasks from 15 websites and challenges agents to ground and reason in highly dynamic and multimodal web environments. We follow WebSight~\cite{bhathal2025websight}, and use a subset of achievable tasks for evaluation.

    \item[(ii)] \textbf{Multimodal-Mind2Web}~\cite{mind2web, zheng2024seeact_multimodal_mind2web} is a large-scale benchmark containing over 2,000 open-ended tasks from 137 real-world websites spanning 31 domains. For our evaluation, we select the first 100 tasks from the test-domain and test-website subsets, which include domains and websites not seen during training, thereby evaluating the out-of-distribution (OOD) generalizability of the agent. Tasks are skipped if the corresponding website is inaccessible at evaluation time.

    \item[(iii)] \textbf{MMInA}~\cite{zhang2024mmina} is designed to evaluate CUAs on real-world websites. It contains 1,050 tasks across domains such as shopping and travel, requiring agents to perform multimodal grounding and long-horizon planning. We use the Wikipedia domain and test on the first 100 samples for evaluation.

    \item[(iv)] \textbf{OSWorld}~\cite{osworld} is a scalable, real computer environment for benchmarking multimodal agents on open-ended tasks across operating systems like Ubuntu, Windows, and macOS. The benchmark includes 369 tasks derived from real-world use cases, covering web and desktop apps, OS file I/O, and workflows spanning multiple applications. It is designed to test broader generalist agent capabilities beyond web-only interaction.
\end{itemize}

For WebVoyager, Multimodal-Mind2Web, and MMInA, we adopt the LLM-as-Judge paradigm~\cite{he2024webvoyager} to evaluate task completion. For MMInA, we provide the language model with both the model's final answer and the ground truth answer, and ask it to determine whether the response is correct. For Multimodal-Mind2Web and WebVoyager, we supply the task description along with a sequence of trajectory screenshots to a VLM, which judges whether the task has been successfully completed. In contrast, OSWorld employs a rule-based evaluation method, using a custom execution-based evaluation script for reliable and reproducible assessment. We report Task Accuracy as the primary evaluation metric across all benchmarks.

\section{Action Space}\label{append:action_space}

The agent operates over a unified action space $\mathcal{A}$ shared between web browsers and desktop operating systems, with small extension sets for platform-specific actions. Each action takes coordinate or string arguments where applicable.

\begin{fancybox}[yellow]{Action Space $\mathcal{A}$}
\small
\begin{tabular}{@{}p{0.42\linewidth}@{\hspace{0.6em}}p{0.50\linewidth}@{}}
\multicolumn{2}{@{}l}{\textcolor{gray!70}{\textit{Base actions: shared by web and desktop OS}}}\\
\texttt{CLICK(x, y)} & click at screen coordinate \\
\texttt{TYPE(text)} & type into focused element \\
\texttt{SCROLL(direction)} & up, down, left, or right \\
\texttt{WAIT(seconds)} & pause for page or app to settle \\
\texttt{KEY\_PRESS(key)} & single key, e.g., Enter or Delete \\
\texttt{STOP(answer)} & terminate task with optional answer \\
\addlinespace[2pt]
\multicolumn{2}{@{}l}{\textcolor{gray!70}{\textit{Web-browser extensions}}}\\
\texttt{NEW\_TAB(url)} & open new tab and navigate \\
\texttt{GOTO\_URL(url)} & navigate active tab to URL \\
\texttt{GO\_BACK()}, \texttt{GO\_FORWARD()} & browser history navigation \\
\texttt{PAGE\_CLOSE()} & close active tab \\
\texttt{SELECT(element\_id, value)} & choose value from dropdown \\
\addlinespace[2pt]
\multicolumn{2}{@{}l}{\textcolor{gray!70}{\textit{Desktop-OS extensions}}}\\
\texttt{DOUBLE\_CLICK(x, y)} & double-click at coordinate \\
\texttt{RIGHT\_CLICK(x, y)} & right-click for context menu \\
\texttt{DRAG\_TO(x1, y1, x2, y2)} & drag between two coordinates \\
\texttt{HOTKEY(keys)} & keyboard shortcut, e.g., Ctrl+C \\
\texttt{MOVE\_TO(x, y)} & position cursor without clicking \\
\end{tabular}
\end{fancybox}

\section{Memory Construction}
\label{append:memory_construct}

The memory bank is constructed from successful trajectories generated via our data flywheel. For our experiments in the web-browser setting, we curate a memory bank of 10,000 trajectories. To ensure diversity, these trajectories are sampled evenly from 13 distinct domains: education, tech, entertainment, travel, health, news, services, shopping, social, food, academic, government, and finance. We apply a quality control process to the selected trajectories: we filter out those with a length shorter than 3 or longer than 15 steps, and we also remove any redundant actions within a trajectory. A more detailed description of our data quality control measures is provided in Appendix~\ref{append:data_quality_control}.

For experiments in the operating system (OS) setting, we employ the GUIAct~\cite{chen2025guiact} dataset as our memory bank. Notably, the original GUIAct dataset provides observations and ground-truth actions but lacks the intermediate reasoning steps required by our agent. To address this, we augment the dataset by using Qwen2.5-VL-7B to generate a plausible reasoning process, conditioned on the original observation and ground-truth action.

\section{Training Setting}
\label{append:training_setting}

\subsection{Training Data}
The composition of our training data is tailored to each experimental setting. For the web-browser checkpoint, we create a composite dataset of 1,410 trajectories, consisting of 1,009 trajectories from the Mind2Web training set~\cite{mind2web} and 401 trajectories from our curated memory bank. For the operating system (OS) checkpoint, we sample 1,500 trajectories from the GUIAct dataset~\cite{chen2025guiact}, specifically filtering for those with a length between 3 and 15 steps.

\subsection{Training Configuration}
We employ a parameter- and data-efficient fine-tuning strategy. Specifically, we apply Low-Rank Adaptation (LoRA) to the memory encoder VLM with a rank of 16. To maintain a lightweight model, the Q-Former is designed with shared parameters across all its layers. The model is trained for a single epoch. This efficient setup allows the entire training process to complete in under 20 hours on a single NVIDIA H100 GPU. To ensure a fair and consistent comparison, this exact training configuration is used for all models and ablation studies presented in this work.

\section{Data Fly-Wheel Quality Control}
\label{append:data_quality_control}

To ensure the integrity and reliability of our data flywheel, we have implemented a comprehensive quality control process that addresses both the performance of trajectories and the implementation of robust ethical safeguards. This multi-layered approach guarantees that the memory bank is populated with high-quality, safe, and effective data.

\subsection{Trajectory Success and Quality Verification}
We employ a three-stage process to verify the quality of each trajectory before its inclusion in our memory bank:

\begin{itemize}
    \item \textbf{LLM-as-Judge:} We utilize a fine-tuned validation model, SEAgent-1.0-7B~\cite{SEAgent}, to automatically evaluate trajectory success. This model is specifically trained for computer-use task verification, providing a reliable, automated first pass on task completion.

    \item \textbf{Rule-Based Filtering:} Following the initial evaluation, we apply a set of deterministic heuristics to further refine the data. This step removes failed trajectories and filters out redundant actions within successful ones. Only successful and non-duplicated steps are ultimately retained for memory construction.

    \item \textbf{Stronger VLM Audit:} To continuously monitor and validate the quality of our memory bank, we conduct periodic audits. In a representative audit, we randomly sampled 100 trajectories and had each one evaluated three times by GPT-4o. The evaluation confirmed a high degree of accuracy, with GPT-4o achieving an average accuracy of 91.3\%.
\end{itemize}

These results indicate that our memory bank maintains high quality and low noise. For full transparency, we provide the audited trajectories and their corresponding evaluation records in the \texttt{trajectory\_evaluation} folder, included in the supplementary materials.

\section{Task Generation Pipeline}
\label{sec:task generation pipeline}

From $\mathcal{Q}_0$, the seed dataset, task descriptions are sampled from each category, of which there are 13 (`education', `tech', `entertainment', `travel', `health', `news', `services', `shopping', `social', `food', `academic', `government', `finance').

Each task is used as a search query in DuckDuckGo using SerpAPI~\cite{serpapi2023}, after which the top 20 search results are collected and added to the dictionary of URLs, under the category that the search query was sampled from, resulting in 6,676 unique page links.

As the first part of the task generation process, each link is processed to truncate them down to the parent domain and first path level. The link is opened with Playwright and a screenshot is taken. The link and screenshot are passed to the VLM with the following prompt to extract the page content and check if the page is inaccessible.

\begin{fancybox}[blue]{Page Content Extraction Prompt}
You are an expert web page analyzer. Analyze the given website and screenshot and extract all useful information in detail. If the page is blocked (\eg 403, 404, has captcha, etc.), directly return `blocked'. Otherwise, return a detailed description of the page content.
\end{fancybox}

The extracted page content then forms part of the prompt to the VLM to extract detailed information on the page tailored specifically for the task generation LLM's use. At this stage, if a page is blocked, we skip it.

\begin{fancybox}[green]{Page Information Extraction Prompt}
You are an expert web page analyzer. Analyze the given website and screenshot and extract all useful information for task generation.
\bigskip

URL: \{url\}

Category: \{category\}

Page Content: \{page\_content\}

\bigskip
Based on the URL, category, and page content above, provide a comprehensive summary of:

1. Website type and purpose - its primary intended use cases, problems that can be solved on this site

2. Navigation structure and key sections

3. Interactive elements (buttons, forms, links, etc.)

4. Content types present (articles, product listings, archives, etc.)

5. Anything unusual or unique to this site

6. IGNORE any cookie notices, privacy popups, log in buttons, site errors, ads, or other elements not specific to the site's function.

\bigskip
Focus on information that would be useful for generating problems for web automation to solve. Think about what human users might want to achieve on this website.

Provide a clear, structured summary that captures these essential aspects of the website for task generation purposes.
\end{fancybox}

The resulting detailed page description forms \{info\_summary\}, part of the task generation prompt. Each prompt also contains \{examples\_section\}, a random selection of 5 example tasks from $\mathcal{Q}_0$, the seed dataset. Examples are pulled from the same task category that the current page belongs to.

\begin{fancybox}[orange]{Initial Task Generation Prompt}
You are an expert generator of task problems for web automation. Based on the website analysis and screenshot, generate 10 diverse task problems that a web agent could solve on this website.

\bigskip
URL: \{url\}

Category: \{category\}

\bigskip
IMPORTANT: Generate direct, actionable problems with a solution. Task problems should be specific and achievable.

\bigskip
Generate exactly 10 diverse, direct instruction task problems that:

1. Have a clear, specific objective; a problem to solve

2. Are achievable within the website's scope

3. Test diverse skills like navigation, information gathering, information synthesis in order to solve problems

4. Require multiple steps to solve, not a single action

5. Have measurable, verifiable success criteria; avoid vague, unverifiable tasks like `Read a paragraph' or `Explore a page'. Instead, focus on a clear, deliberate end goal.

6. IMPORTANT: Do NOT write tasks that contain overly specific instructions like `Click on X...' or `Type X...'. These are not problems to be solved. The task problem should not direct the agent on what to do, but rather what to achieve.

7. Are distinct, not related to each other, and do not require knowledge or completion of previous task problems.

\bigskip
For each task problem, provide:

- Task problem description (direct instruction with specific requirements)

- Expected outcome (what should be accomplished, what condition needs to be checked to indicate success)

- Difficulty level (easy/medium/hard)

\bigskip

\{examples\_section\}

\bigskip
Format your response as exactly 10 tasks, one per line, with this structure:

1. [Task Description] \textbar{} [Expected Outcome] \textbar{} [Difficulty]

2. [Task Description] \textbar{} [Expected Outcome] \textbar{} [Difficulty]

3. [Task Description] \textbar{} [Expected Outcome] \textbar{} [Difficulty]

\bigskip
Example format:

1. Find a news article about climate change published in the last week \textbar{} Should locate and display a recent climate change article \textbar{} Easy

2. Search for iPhone 12 Pro with price below \$800 \textbar{} Should have navigated to a page for iPhone 12 Pro models under \$800 \textbar{} Medium

3. Book a hotel in New York for 2 people for next weekend \textbar{} Should find and select a specific hotel for 2 people \textbar{} Medium

4. Find the top 3 customer reviews for Nike running shoes \textbar{} Should locate and display a page showing the top 3 reviews \textbar{} Easy

5. Compare prices of Samsung Galaxy phones between \$500-700 \textbar{} Should find a page comparing multiple Samsung phones in that price range \textbar{} Medium

6. Find the paper with the most citations under the Computer Science category \textbar{} Should identify and display the page of an AI paper \textbar{} Hard

\bigskip
Respond only with the 10 tasks in the specified format, no additional text.

\bigskip
Website Analysis: \{info\_summary\}
\end{fancybox}

The tasks generated by the LLM in this first pass contain some poor-quality tasks that are overly specific and not similar to tasks undertaken by human users.

To refine these tasks, each generated task is passed to the LLM again with a prompt to rewrite the task if necessary. The prompt is based on the one used in ZeroGUI's~\cite{ZeroGUI} new task generation pipeline.

\begin{fancybox}[purple]{Task Refinement Prompt}
You will be given a task instruction. Please refine it with the following requirements:
\bigskip

1. Imitate the speech style of a human user. Make it more natural and diverse.

2. Remove specific hints on how to achieve the task. (\eg press which button, click which link, etc.)

3. The task should remain unambiguous and clear, and the task's goal must remain the same.
\bigskip

For example, you should refine:

``Use the `Price range' filter on the Ryanair website to limit the flight options to \$50-100.''

into:

``Find flights from London to Paris with a price range of \$50-100.''

or

``I only have a limited budget for flights. Could you help find flights from London to Paris with a price range of \$50-100.''
\bigskip

You should refine:

``Import email messages from another email program by clicking the `Import' button under the `Import from Another Program' section.''

into:

``Please import email messages from another email program.''

or

``I have some emails in my other email program. Could you help me import them into Thunderbird?''
\bigskip

You should refine:

``Click on the `About Us' link to learn more about the company's history and mission.''

into:

``Please provide information about the company's history and mission.''

or

``Find an `About Us' or similar page on the website that describes the company's history and mission.''
\bigskip

The original instruction is:

\{instruction\}

Refine it and return the refined instruction text in this exact format:
\bigskip

ORIGINAL: [the original instruction]

REFINED: [the refined instruction]
\end{fancybox}

\vspace{1cm}

\section{Broader Impacts}
\label{append:broader_impacts}

\paragraph{Positive impacts.}
Continuous-memory-augmented GUI agents reduce the cost and latency of automating routine, multi-step computer-use tasks. This has direct applications in accessibility, where users with motor or visual impairments can delegate repetitive interaction sequences to a reliable assistant; in workplace productivity, where common forms, lookups, and software workflows can be automated; and in digital education, where agents demonstrate procedural skills across applications. The autonomous data flywheel further reduces the human-annotation burden of training such agents, broadening access to high-quality computer-use data for the research community.

\paragraph{Ethical and Robustness Safeguards.}
Beyond performance metrics, we have integrated three layers of ethical control throughout our data pipeline to ensure safe and responsible agent operation:

\begin{itemize}
    \item \textbf{Website Collection:} Data collection is initiated using SerpAPI~\cite{serpapi2023}, which offers legal protections and indexes only publicly accessible content. To ensure relevance and safety, seed queries for web exploration are sourced from established public datasets.

    \item \textbf{Task Generation:} We use an LLM as a filter to prevent the generation of unsafe or ethically sensitive tasks. This safeguard ensures that the agent does not attempt to perform inappropriate or harmful interactions.

    \item \textbf{Task Execution:} The agent is designed to be respectful of website policies. If a website blocks access or if the agent's automated behavior is detected, it immediately halts execution. This mechanism prevents the scraping of protected content and respects the operational rules of web domains.
\end{itemize}

We are committed to upholding high ethical standards and plan to enhance these safeguards in future work by incorporating freshness tracking, more extensive deduplication audits, and advanced privacy filters.

\section{Limitations}
\label{append:limitations}

Our backbone-scaling experiment in Section~\ref{sec:scal_backbone} mixes two Qwen-VL families: we use Qwen3-VL~\cite{yang2025qwen3} for the 2B/4B/8B points and Qwen2.5-VL~\cite{yang2024qwen2.5} for the 32B point. The reason is that Qwen3-VL's release at the 30B regime is a mixture-of-experts variant whose effective inference compute is not directly comparable to the dense 2B/4B/8B points. We therefore use the dense Qwen2.5-VL-32B to keep the architecture-consistent sweep, at the cost of a single-family controlled experiment.

Second, our scaling study is bounded by training resources. The largest backbone we evaluate is Qwen2.5-VL-32B; extending to the 200B-class regime, such as Qwen3-VL-235B, would require compute resources beyond what was available to us for memory encoder training. While the evaluated 2B--32B backbones show positive empirical trends as described in Section~\ref{sec:scaling_laws}, validating this trend at substantially larger scales is left for future work with greater compute resources.

\section{Case Study}
\subsection{Shopping Scenario}
\begin{figure}[h]
    \centering
    \includegraphics[width=\linewidth]{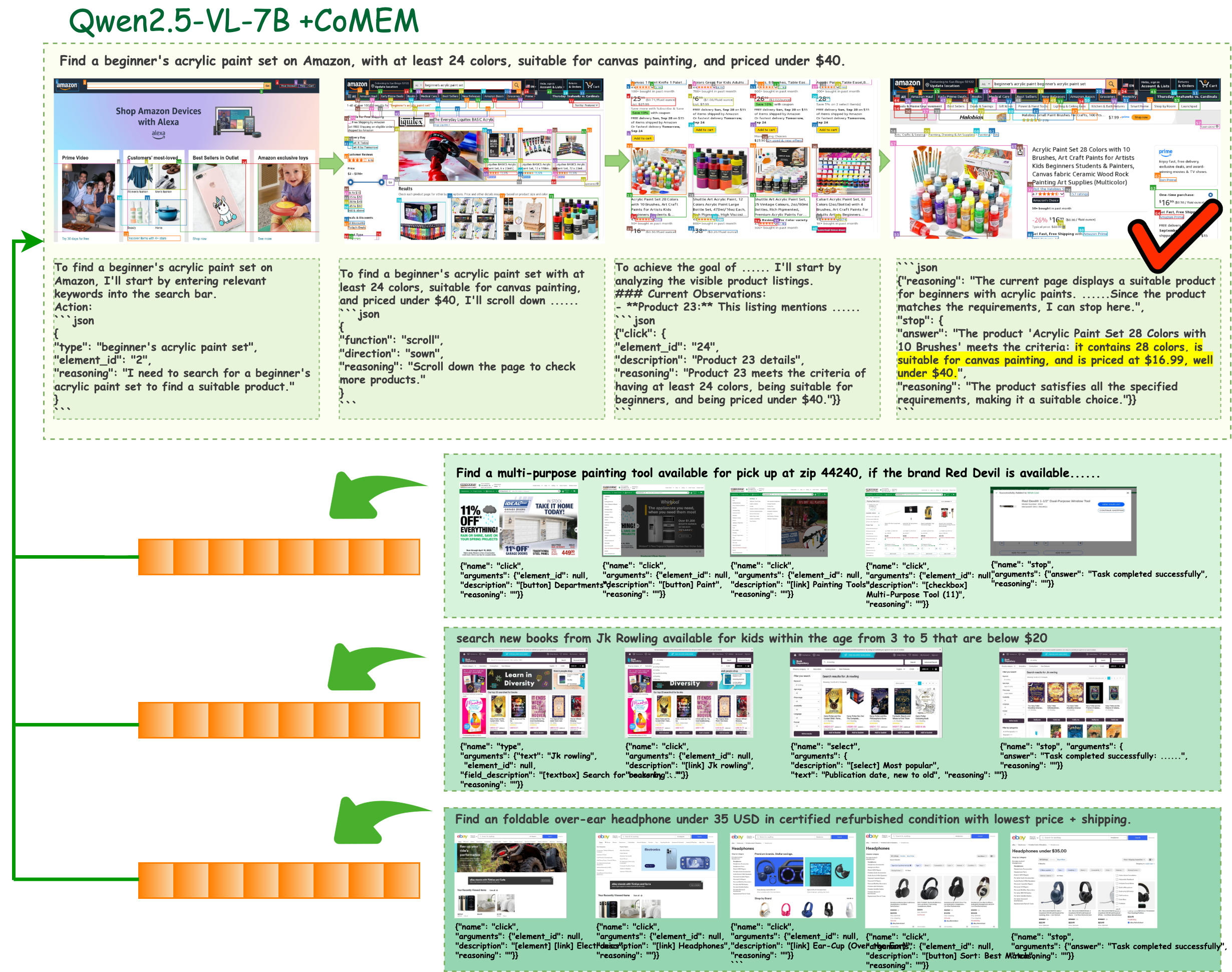}
\caption{Successful case 1 by Qwen2.5-VL-7B + Ours in shopping scenario. The model accurately understands the goal, retrieves relevant memory, and selects the correct product.}
    \label{fig:succes_1}
\end{figure}

\begin{figure}[h]
    \centering
    \includegraphics[width=\linewidth]{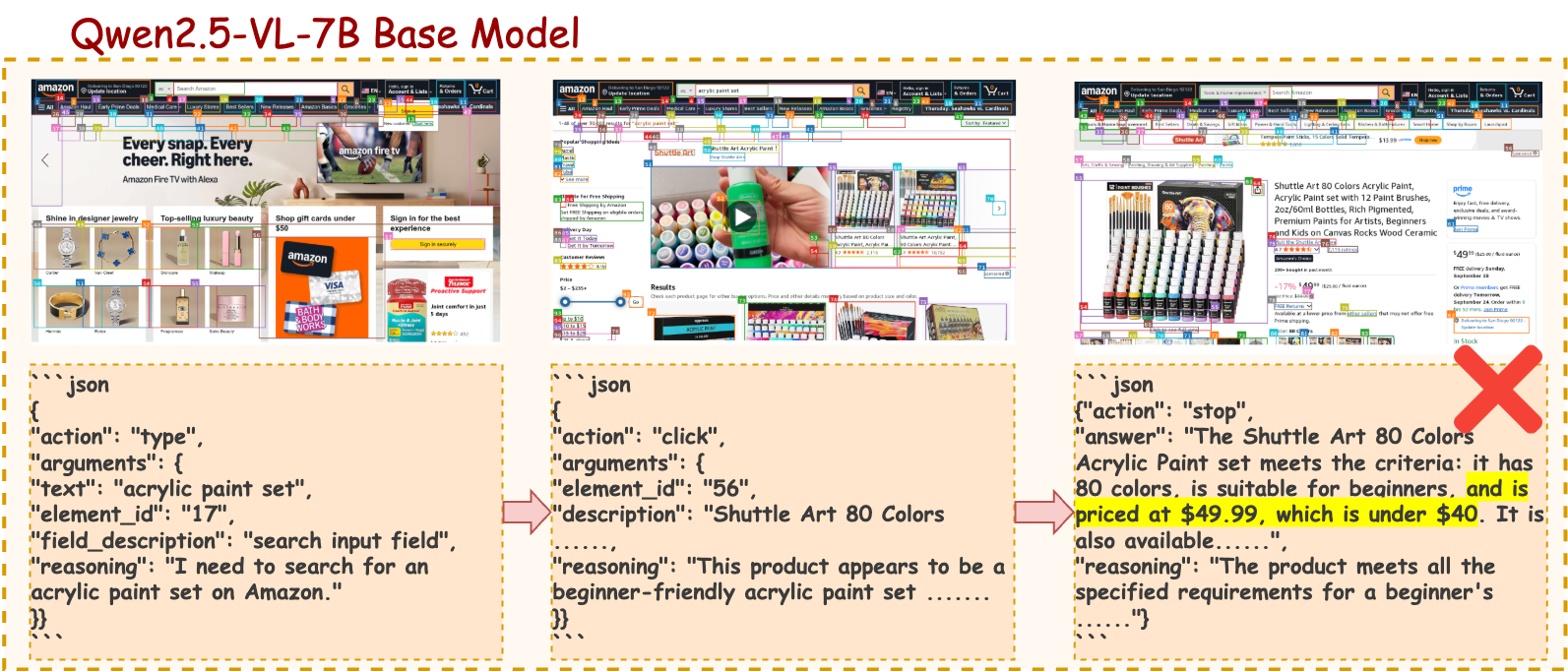}
\caption{Failure case of the base model Qwen2.5-VL-7B without memory.
The model selects an incorrect product that exceeds the price limit, highlighting the importance of contextual memory.}
    \label{fig:fail_1}
\end{figure}

In the task ``Find a beginner's acrylic paint set on Amazon, with at least 24 colors, suitable for canvas painting, and priced under \$40,'' the base model exhibits limited reasoning and planning capabilities as shown in Figure~\ref{fig:fail_1}. It prematurely selects the first product shown, failing to rigorously evaluate whether the item meets all specified constraints such as price, color count, and suitability for canvas painting. This behavior suggests a lack of contextual understanding and strategic planning. In contrast, the Ours method demonstrates significantly improved performance by leveraging compressed memory representations from multiple prior shopping trajectories, as shown in Figure~\ref{fig:succes_1}. These compact embeddings provide rich context, guiding the model to follow a more deliberate and accurate plan, searching, analyzing multiple candidate products, and validating them against all task constraints before making a final selection. This example highlights how continuous memory enables deeper reasoning and constraint-aware decision-making, leading to more reliable outcomes in complex multi-step tasks.

\subsection{Information Scenario}

\begin{figure}[h]
    \centering
    \includegraphics[width=\linewidth]{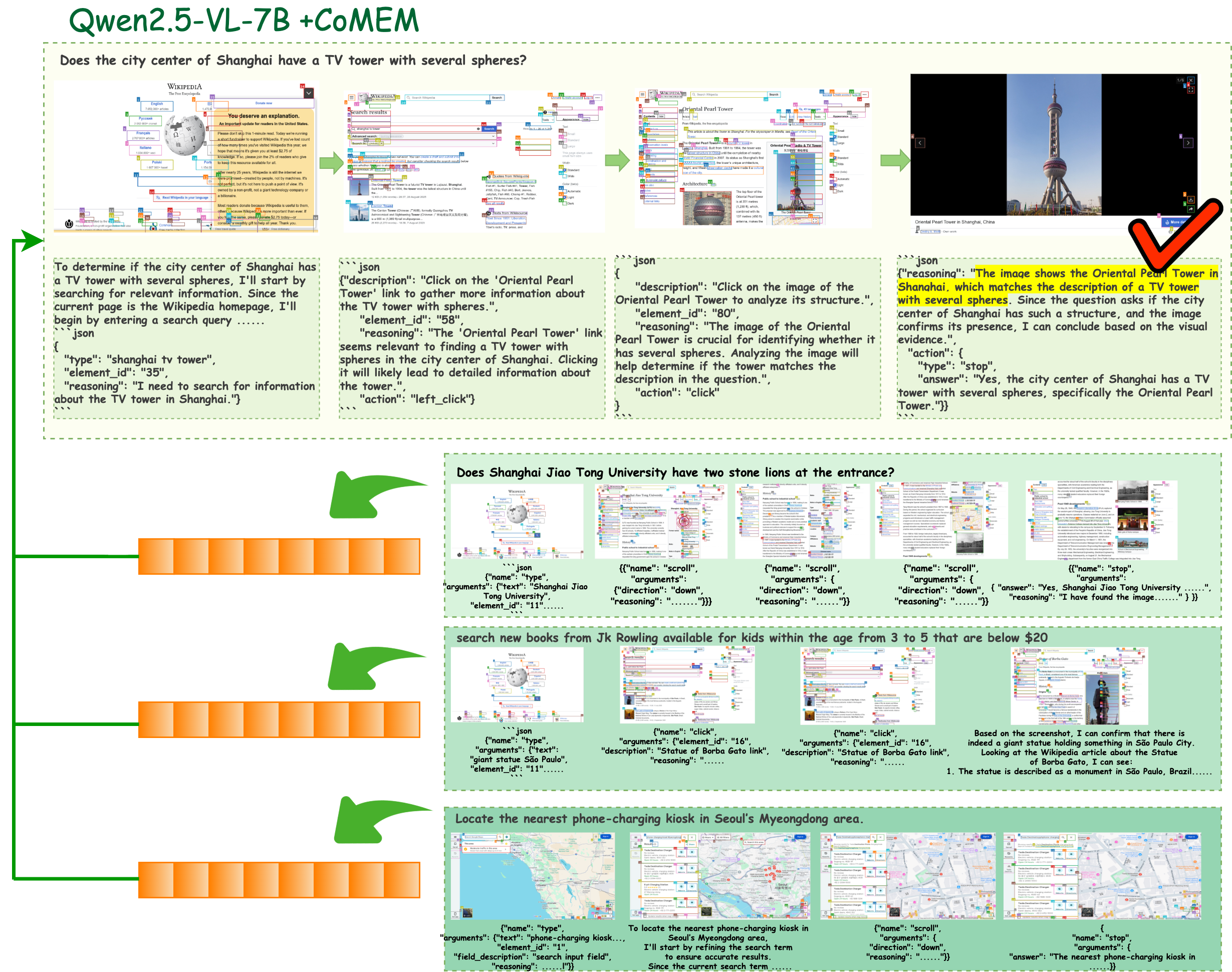}
\caption{Successful case 2 by Qwen2.5-VL-7B + Ours in information scenario. The model accurately understands the goal, retrieves relevant memory, and finds the answer.}
    \label{fig:success_2}
\end{figure}

\begin{figure}[h]
    \centering
    \includegraphics[width=\linewidth]{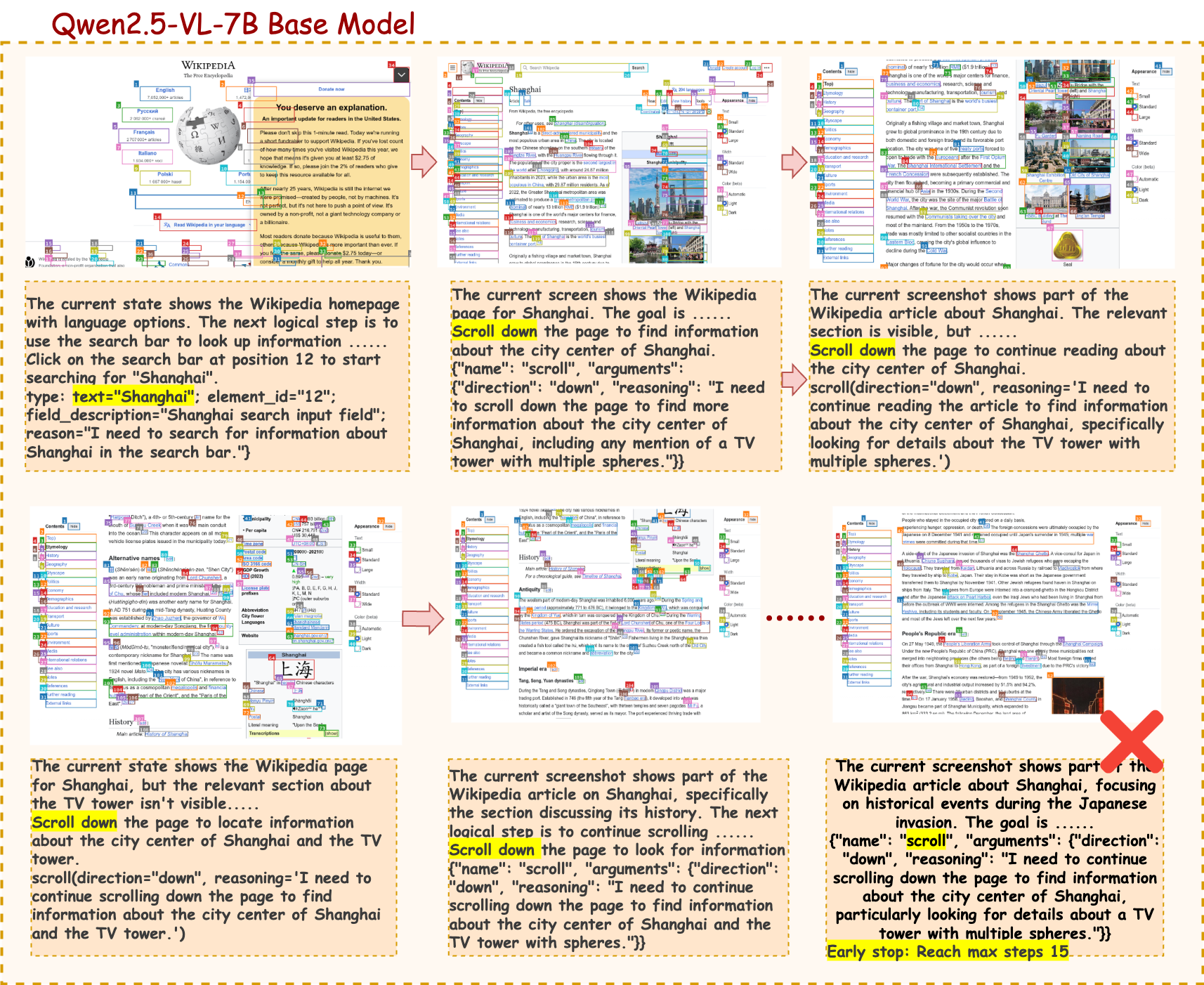}
\caption{Failure case of the base model Qwen2.5-VL-7B without memory. The model starts with a too general query and lacks enough exploration.}
    \label{fig:fail_2}
\end{figure}

In the task ``Does the city center of Shanghai have a TV tower with several spheres?'', the base model demonstrates inefficient behavior by issuing an overly general query (``Shanghai'') and repeatedly scrolling through irrelevant sections of the Wikipedia article. As shown in Figure~\ref{fig:fail_2}, this lack of targeted exploration causes the model to miss the visual and textual cues needed to answer the question, ultimately resulting in failure. In contrast, in Figure~\ref{fig:success_2} the Ours method begins with a more precise search query (``Shanghai TV tower'') informed by prior experience. It quickly identifies the Oriental Pearl Tower, whose structure features multiple spheres, and successfully verifies its location using both textual and visual evidence. This comparison highlights how Ours enables more focused planning and efficient information retrieval, especially in tasks requiring multimodal reasoning.

\clearpage

\newpage
\input{checklist.tex}

\end{document}

%% file: checklist.tex
\section*{NeurIPS Paper Checklist}

\begin{enumerate}

\item {\bf Claims}
    \item[] Question: Do the main claims made in the abstract and introduction accurately reflect the paper's contributions and scope?
    \item[] Answer: \answerYes{}
    \item[] Justification: The abstract and introduction state our three main contributions, thoroughly explore the design space for continuous memory architecture and its scaling study, the autonomous data flywheel, and validation on computer-use agent benchmarks. Each contribution is supported by corresponding sections.

\item {\bf Limitations}
    \item[] Question: Does the paper discuss the limitations of the work performed by the authors?
    \item[] Answer: \answerYes{}
    \item[] Justification: A dedicated Limitations section in the appendix discusses the backbone-family mix and the compute-bounded scale ceiling of our experiments.

\item {\bf Theory assumptions and proofs}
    \item[] Question: For each theoretical result, does the paper provide the full set of assumptions and a complete (and correct) proof?
    \item[] Answer: \answerNA{}
    \item[] Justification: This is an empirical study, no formal theoretical results are claimed.

    \item {\bf Experimental result reproducibility}
    \item[] Question: Does the paper fully disclose all the information needed to reproduce the main experimental results of the paper to the extent that it affects the main claims and/or conclusions of the paper (regardless of whether the code and data are provided or not)?
    \item[] Answer: \answerYes{}
    \item[] Justification: Section 4.1 and the appendices on Memory Construction, Training Setting, and Data Fly-Wheel Quality Control document model identities, hyperparameters, training procedures, and benchmark protocols.

\item {\bf Open access to data and code}
    \item[] Question: Does the paper provide open access to the data and code, with sufficient instructions to faithfully reproduce the main experimental results, as described in supplemental material?
    \item[] Answer: \answerYes{}
    \item[] Justification: We will fully open-source the trajectory dataset, the data flywheel pipeline, inference code and the memory-encoder training code upon acceptance.

\item {\bf Experimental setting/details}
    \item[] Question: Does the paper specify all the training and test details (e.g., data splits, hyperparameters, how they were chosen, type of optimizer) necessary to understand the results?
    \item[] Answer: \answerYes{}
    \item[] Justification: Section 4.1, titled Experimental Setup, and the appendix on Training Setting specify data splits, hyperparameters, and evaluation protocols.

\item {\bf Experiment statistical significance}
    \item[] Question: Does the paper report error bars suitably and correctly defined or other appropriate information about the statistical significance of the experiments?
    \item[] Answer: \answerNo{}
    \item[] Justification: Following standard practice for computer-use agent benchmarks, results are reported as success rates over the fixed test sets without error bars; running across five backbones, nine web-domain splits, and ten OS-suite splits was not feasible within our compute budget.

\item {\bf Experiments compute resources}
    \item[] Question: For each experiment, does the paper provide sufficient information on the computer resources (type of compute workers, memory, time of execution) needed to reproduce the experiments?
    \item[] Answer: \answerYes{}
    \item[] Justification: We disclose the data collection cost of \$1{,}972 USD in Section 5, and the lightweight LoRA + Q-Former training configuration tuning 1.2\% of parameters on 1.5k samples in the appendix on Training Setting, alongside the model identities used throughout the paper.
    
\item {\bf Code of ethics}
    \item[] Question: Does the research conducted in the paper conform, in every respect, with the NeurIPS Code of Ethics \url{https://neurips.cc/public/EthicsGuidelines}?
    \item[] Answer: \answerYes{}
    \item[] Justification: The work conforms to the NeurIPS Code of Ethics; the data flywheel includes content-safety filters described in the appendix on Ethical and Robustness Safeguards.

\item {\bf Broader impacts}
    \item[] Question: Does the paper discuss both potential positive societal impacts and negative societal impacts of the work performed?
    \item[] Answer: \answerYes{}
    \item[] Justification: Positive impacts include lowering the cost of automating routine multi-step computer-use tasks, with applications in accessibility and productivity. Ethical and Robustness Safeguards about data collection and agent safety are also discussed.
    
\item {\bf Safeguards}
    \item[] Question: Does the paper describe safeguards that have been put in place for responsible release of data or models that have a high risk for misuse (e.g., pre-trained language models, image generators, or scraped datasets)?
    \item[] Answer: \answerYes{}
    \item[] Justification: The appendix on Ethical and Robustness Safeguards describes the three-layer content-filtering pipeline applied during automatic trajectory collection, covering website selection, task generation, and task execution.

\item {\bf Licenses for existing assets}
    \item[] Question: Are the creators or original owners of assets (e.g., code, data, models), used in the paper, properly credited and are the license and terms of use explicitly mentioned and properly respected?
    \item[] Answer: \answerYes{}
    \item[] Justification: All third-party models, datasets, benchmarks, and tools, such as Qwen-VL, GPT-4o, Mind2Web, OSWorld, GUIAct, and SerpAPI, are cited at first use; we follow their respective licenses and terms of use.

\item {\bf New assets}
    \item[] Question: Are new assets introduced in the paper well documented and is the documentation provided alongside the assets?
    \item[] Answer: \answerYes{}
    \item[] Justification: The 188{,}451-trajectory dataset and the data flywheel pipeline introduced in Section 5 will be fully open-sourced upon acceptance, with documentation provided in the appendices on Task Generation Pipeline and Data Fly-Wheel Quality Control.

\item {\bf Crowdsourcing and research with human subjects}
    \item[] Question: For crowdsourcing experiments and research with human subjects, does the paper include the full text of instructions given to participants and screenshots, if applicable, as well as details about compensation (if any)?
    \item[] Answer: \answerNA{}
    \item[] Justification: The data collection pipeline is fully automated using VLMs as task generators and trajectory judges; no crowdsourcing or human subjects are involved.

\item {\bf Institutional review board (IRB) approvals or equivalent for research with human subjects}
    \item[] Question: Does the paper describe potential risks incurred by study participants, whether such risks were disclosed to the subjects, and whether Institutional Review Board (IRB) approvals (or an equivalent approval/review based on the requirements of your country or institution) were obtained?
    \item[] Answer: \answerNA{}
    \item[] Justification: This work does not involve human subjects or crowdsourced data collection; IRB approval is therefore not applicable.

\item {\bf Declaration of LLM usage}
    \item[] Question: Does the paper describe the usage of LLMs if it is an important, original, or non-standard component of the core methods in this research? Note that if the LLM is used only for writing, editing, or formatting purposes and does \emph{not} impact the core methodology, scientific rigor, or originality of the research, declaration is not required.
    \item[] Answer: \answerYes{}
    \item[] Justification: VLMs are integral to this work as agent backbones in Section 4, as the encoder of continuous memory in Section 2, and as the task generator and trajectory judge in the data flywheel in Section 5.

\end{enumerate}